\pdfoutput=1

\documentclass[11pt]{article}

\usepackage{EMNLP2022}

\usepackage{times}
\usepackage{latexsym}

\usepackage[T1]{fontenc}

\usepackage[utf8]{inputenc}

\usepackage{microtype}

\usepackage{inconsolata}

\usepackage{graphicx}
\usepackage{pdfpages}
\usepackage{url}
\usepackage{hyperref}
\usepackage{xcolor}
\usepackage{epsfig}
\usepackage{adjustbox}
\usepackage{amsfonts}
\usepackage{amsmath}
\usepackage{amssymb}
\usepackage{booktabs} 
\usepackage{comment}
\usepackage{multirow}
\usepackage{caption}
\usepackage{subcaption}
\usepackage{textcomp}
\usepackage{relsize}
\usepackage{stmaryrd}
\usepackage{bbm}
\usepackage{rotating}
\usepackage{helvet}
\usepackage{courier}
\usepackage{natbib}
\usepackage{hyperref}

\title{\textsc{BotsTalk}: Machine-sourced Framework for Automatic Curation of Large-scale Multi-skill Dialogue Datasets}

\author{
        Minju Kim\textsuperscript{\rm 1}\thanks{$^\ast$Equal contribution}~~~
        Chaehyeong Kim\textsuperscript{\rm 1}$^\ast$~~~
        Yongho Song\textsuperscript{\rm 1$\ast$}~~~
        Seung-won Hwang\textsuperscript{\rm 2}~~~
        Jinyoung Yeo\textsuperscript{\rm 1}\thanks{$^\dagger$Corresponding author}\\
        \textsuperscript{\rm 1}Department of Artificial Intelligence, Yonsei University\\
        \textsuperscript{\rm 2}Department of Computer Science and Engineering, Seoul National University\\
        \texttt{\{minnju,cheris8,kopf\_yhs,jinyeo\}@yonsei.ac.kr}~~~\texttt{seungwonh@snu.ac.kr}
        }

\begin{document}
\maketitle

\newcommand{\se}{{\it SE}}
\newcommand{\eg}{{\it e.g.}}
\newcommand{\ie}{{\it i.e.}}
\newcommand{\etal}{{\it et al.}}
\newcommand{\etc}{{\it etc}}

\newcommand{\mcal}[1]{{\cal{#1}}}
\newcommand{\calA}{\mbox{${\cal A}$}}
\newcommand{\calB}{\mbox{${\cal B}$}}
\newcommand{\calC}{\mbox{${\cal C}$}}
\newcommand{\calD}{\mbox{${\cal D}$}}
\newcommand{\calE}{\mbox{${\cal E}$}}
\newcommand{\calF}{\mbox{${\cal F}$}}
\newcommand{\calG}{\mbox{${\cal G}$}}
\newcommand{\calH}{\mbox{${\cal H}$}}
\newcommand{\calI}{\mbox{${\cal I}$}}
\newcommand{\calJ}{\mbox{${\cal J}$}}
\newcommand{\calK}{\mbox{${\cal K}$}}
\newcommand{\calL}{\mbox{${\cal L}$}}
\newcommand{\calM}{\mbox{${\cal M}$}}
\newcommand{\calN}{\mbox{${\cal N}$}}
\newcommand{\calO}{\mbox{${\cal O}$}}
\newcommand{\calP}{\mbox{${\cal P}$}}
\newcommand{\calQ}{\mbox{${\cal Q}$}}
\newcommand{\calR}{\mbox{${\cal R}$}}
\newcommand{\calS}{\mbox{${\cal S}$}}
\newcommand{\calT}{\mbox{${\cal T}$}}
\newcommand{\calU}{\mbox{${\cal U}$}}
\newcommand{\calV}{\mbox{${\cal V}$}}
\newcommand{\calW}{\mbox{${\cal W}$}}
\newcommand{\calX}{\mbox{${\cal X}$}}
\newcommand{\calY}{\mbox{${\cal Y}$}}
\newcommand{\calZ}{\mbox{${\cal Z}$}}

\newcommand*\concat{\mathbin{\|}}

\makeatletter
\newcommand*\Neg[2][0mu]{\Neginternal{#1}{\negslash}{#2}}
\newcommand*\sNeg[2][0mu]{\Neginternal{#1}{\snegslash}{#2}}
\newcommand*\ssNeg[2][0mu]{\Neginternal{#1}{\ssnegslash}{#2}}
\newcommand*\sssNeg[2][0mu]{\Neginternal{#1}{\sssnegslash}{#2}}
\newcommand*\Neginternal[3]{\mathpalette\Neg@{{#1}{#2}{#3}}}
\newcommand*\Neg@[2]{\Neg@@{#1}#2}
\newcommand*\Neg@@[4]{%
  \mathrel{\ooalign{%
    $\m@th#1#4$\cr
    \hidewidth$\m@th#3{#1}\mkern\muexpr#2*2$\hidewidth\cr
  }}%
}

\newcommand*\negslash[1]{\m@th#1\not\mathrel{\phantom{=}}}
\newcommand*\snegslash[1]{\rotatebox[origin=c]{60}{$\m@th#1-$}}
\newcommand*\ssnegslash[1]{\rotatebox[origin=c]{60}{$\m@th#1{\dabar@}\mkern-7mu{\dabar@}$}}
\newcommand*\sssnegslash[1]{\rotatebox[origin=c]{60}{$\m@th#1\dabar@$}}

\newcommand{\aggregate}[2]{\underset{#2}{\operatornamewithlimits{#1\ }}}
\newcommand{\agent}[0]{\includegraphics[width=.03\textwidth]{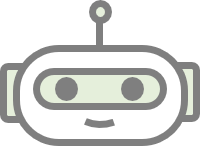}}
\newcommand{\moderator}[0]{\includegraphics[width=.021\textwidth]{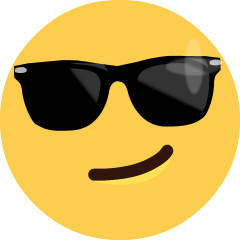}}

\definecolor{lightblue}{RGB}{224,236,247}
\definecolor{deepblue}{RGB}{9,46,107}
\begin{abstract}
To build open-domain chatbots that are able to use diverse communicative skills, we propose a novel framework \textsc{BotsTalk}, where multiple agents grounded to the specific target skills participate in a conversation to automatically annotate multi-skill dialogues. We further present Blended Skill BotsTalk (BS$\mathbb{B}$T), a large-scale multi-skill dialogue dataset comprising 300K conversations. Through extensive experiments, we demonstrate that our dataset can be effective for multi-skill dialogue systems which require an understanding of skill blending as well as skill grounding. Our code and data are available at \url{https://github.com/convei-lab/BotsTalk}.
\end{abstract}
\section{Introduction}

A considerable progress has been made towards open-domain chatbots with different desirable qualities in conversations. Each of these models is capable of being specialized in one communicative skill, \ie, skill grounding. A number of distinct large-scale datasets targeting a specific conversational skill have recently become available. 
ConvAI2~\citep{dinan2019convai2} is provided for research work that aims to endow chatbots with personas~\citep{majumder2020like,kim2020will}, enabling chatbots to talk about themselves. 
Wizard of Wikipedia (WoW)~\citep{dinan2019wizard} is a popular option for recent studies~\citep{lian2019learning,zhao2020knowledge,kim2020sequential} that focus on knowledgeable conversational agents discussing topics in depth. Empathetic Dialogues (ED)~\citep{rashkin2019ed} is also commonly used to embody empathy in dialogue systems~\citep{santhanam2019emotional,majumder2020mime}. Most of such skill-grounded datasets are designed to improve a single skill, and thus effective when models are asked to demonstrate the targeted conversational skill.

Benefiting from the advances of these conversational agents, recent research focuses on another aspect of open-domain chatbots: the ability to blend various conversational skills into one cohesive flow in a seamless manner, \ie, skill blending. A good open-domain chatbot should be able to weave multiple behaviors and skills in a single conversation, so that it enables to deal with different users and situations appropriately~\citep{shuster2020dodeca,roller2021recipes}. Towards this goal, there is a need to construct a multi-skill dialogue dataset, which consists of multi-turn dialogues that exhibit multiple skills. While \citet{smith2020bst} propose a crowdsourced dataset Blended Skill Talk (BST) of 5K conversations as a reliable benchmark for measuring dialogue systems' ability at the blended objective, it is not sufficient to build a multi-skill chatbot due to its limited scale. Scaling up crowdsourcing is not feasible, as it requires labor intensive manual annotation and verification. Instead, automatic curation shows promising results on large-scale dialogue generation~\citep{mohapatra2021simulated}.

In this paper, we aim to generate a large-scale multi-skill dialogue dataset without additional costs or human efforts. To this end, we introduce an automatic data curation approach named \textbf{\textsc{BotsTalk}}, where multiple dialogue agents grounded to individual skills engage in the conversation to blend all skills together. Based on this framework, we create \textbf{Blended Skill BotsTalk (BS$\mathbb{B}$T)}, a large-scale multi-skill dialogue dataset of 300K conversations blended and grounded with a number of skills derived from ConvAI2, WoW, and ED. Our experiments demonstrate that by using our dataset dialogue models successfully yield large performance gains in skill blending while maintaining competitive performance in skill grounding. Furthermore, we validate the quality of BS$\mathbb{B}$T dataset by human evaluation, showing our machine-sourced conversations are consistently preferred over crowdsourced ones from BST by human judges  across all metrics.

\begin{table*}[t!]
\begin{center}\small
{
\begin{tabular}{l l}
    \toprule
    \textbf{Dataset} & \textbf{Dialogue episode} \\
    \midrule
    
    \multirow{5}{*}{ConvAI2} 
    & \textbf{Skill context} for speaker A: I like to ski; I hate Mexican food; I like to eat cheetos; ... \\
    & \textbf{Skill context} for speaker B: I am an artist; I have four children; I enjoy walking for exercise; ... \\
    & \textbf{Dialogue context} \\
    & A: How old are your children? \\ 
    & B: I have four that range in age from 10 to 21. You? \\
    \midrule
    
    \multirow{5}{*}{Wizard of Wikipedia} 
    & \textbf{Skill context} for speaker A: Armadillo \\
    & \textbf{Skill context} for speaker B: Armadillo are ... "armadillo" means "little armoured one" in ...\\ 
    & \textbf{Dialogue context} \\
    & A: I don't think I've ever seen an armadillo in real life! \\ 
    & B: I've seen them at the zoo. Armadillo means little armored one in Spanish. \\ 
    \midrule

    \multirow{5}{*}{Empathetic Dialogues} 
    & \textbf{Skill context} for speaker A: My brother jump scared me while I was out playing; Terrified \\ 
    & \textbf{Skill context} for speaker B: None \\ 
    & \textbf{Dialogue context} \\
    & A: Just got scared to death. \\ 
    & B: Oh no. What happened? \\
    
    \bottomrule
\end{tabular}}
\end{center}
\caption{Example dialogues of three single-skill datasets: ConvAI2 provides each speaker persona sentences as skill context;  Wizard of Wikipedia provides a topic and knowledge resources as skill context; Empathetic Dialogues provides a situation description and emotion as skill context. We only provide two turns of dialogue contexts due to the limit on the paper length.}
\label{tab:dialogue_example}
\end{table*}


\section{Related Work}

\subsection{Skill-grounded Dialogue Datasets}

Past research in open-domain chatbots has made solid strides towards dialogue systems with desirable general qualities in a conversation. Generating responses grounded to specific conversational skill has been explored in different axes, as shown in Table~\ref{tab:dialogue_example} (see also Appendix~\ref{sec:single} for details). \citet{dinan2019convai2} introduce ConvAI2 dataset which consists of more than 140K utterances of crowdsourced conversations to make chit-chat models more engaging and personalized by conditioning the models on profile information. Wizard of Wikipedia~\citep{dinan2019wizard} task aims to explore conversation informed by expert knowledge from Wikipedia and provides about 194K utterances of conversations on about 1,250 topics. \citet{rashkin2019ed} construct a dataset, Empathetic Dialogues, comprising 50K utterances of crowdworker conversations grounded in an emotional situation for a model to converse with empathy. However, it remains unclear whether models optimized for performance along specific conversational skill can retain the learned skill while blending it with other skills.

Hence, \citet{smith2020bst} aim to build a conversational agent who seamlessly blends being personable, knowledgeable, and empathetic. In order to gauge how successful a model is at this blended objective, \citet{smith2020bst} collect a new multi-skill dialogue dataset of about 5K conversations, Blended Skill Talk, via crowdsourcing. While this work provides a testbed for future studies, the scale of data could hinder further progress, since training multi-skill chatbots generally requires a large-scale dataset consisting of conversations that involve multiple skills~\citep{shah2018selfplay}.

\subsection{Automatic Dialogue Data Annotation}

Research in dialogue systems has been consistently supported by the development of new dialogue datasets~\citep{williams2014dstc,mrkvsic2016woz}. One popular approach is to collect and annotate dialogues via crowdsourcing~\citep{zhang2018persona,smith2020bst}. However, generating multi-turn dialogues in this manner requires expensive and exhausting human efforts~\citep{shah2018selfplay,sun2020adding,mohapatra2021simulated}.

Therefore, recent study seeks to facilitate open-domain chatbot development with new datasets automatically constructed by using existing datasets. For instance, \citet{lee2021constructingmultimodal} create a 45K multi-modal dialogue dataset by replacing parts of source dialogues from existing text-only dialogue datasets with their semantically relevant images. \citet{sun2020adding} propose a Human $\leftrightarrow$ AI collaborative data collection approach for generating diverse chit-chat response to augment task-oriented dialogues and present new chit-chat based annotations to 23.8K dialogues from two popular task-oriented datasets. \citet{kim2021neuralwoz} and \citet{vidgen2021worst} present a model-based dialogue collection framework and a human-and-model-in-the-loop process for generating datasets respectively. 


\section{Problem Formulation}

In this section, we formulate the problem of multi-skill dialogue annotation and desirable characteristics for the dialogue dataset as a training resource.

\subsection{Multi-skill Dialogue Annotation}

Our goal is to collect a new large-scale multi-skill dialogue dataset, which seamlessly blends various skills over the course of a multi-turn conversation. Here, inspired by \citet{smith2020bst}, the inputs of this task are single-skill datasets, which are separately collected on a variety of skills. Let $\mathbb{M}$ be the set of $M$ skill types, \eg, $\mathbb{M} = \{ \text{P}, \text{K}, \text{E} \}$, where P, K, E denote personality, knowledge, and empathy derived from ConvAI2, WoW, and ED, respectively. Formally, we refer to $\calD_{m}$ as a dialogue dataset with $N_{m}$ dialogue episodes for skill $m \in \mathbb{M}$
\begin{equation}\label{eq:dataset}\small
    \calD_{m}= \{ ( {stx}_{i,m}, {dtx}_{i,t} ) \}_{i=1}^{N_m}
\end{equation}
where $stx_{i,m}$ is a skill-relevant description (\ie, skill context) for skill $m$ and $dtx_{i,t}$ is $t$ dialogue turns (\ie, dialogue context) derived from the skill context, as shown in Table~\ref{tab:dialogue_example}. Based on the input datasets $\calD_1,...,\calD_M$, we aim to obtain a new dialogue dataset $\tilde{\calD}$ for $M$ skills as an output. Formally,
\begin{equation}\small
    \tilde{\calD} = \{ ( \tilde{stx}_{i}, dtx_{i,t} ) \}_{i=1}^{\infty}
\end{equation}
where $\tilde{stx}_{i}$ is a set of skill contexts for $\mathbb{M}$ and $dtx_{i,t}$ is the dialogue context derived from the multiple skills. We will omit the index $i$ when dealing with a single dialogue episode.

\subsection{Desirable Characteristics of Multi-skill Dialogue Datasets}

By the above annotation, we aim to build a multi-skill chatbot that uses all target skills appropriately in a conversation. For that, we lay out two criteria that a multi-skill dialogue dataset should meet as a training resource, namely \textbf{skill blending} and \textbf{skill grounding}. Skill blending indicates that a multi-skill dialogue dataset should enable dialogue models to exhibit different dialogue skills in a conversation~\citep{smith2020bst}, while skill grounding emphasizes that dialogue models should learn to maintain each dialogue skill when appropriate~\citep{shazeer2017outrageously}. Generally, they have a trade-off relationship as it is insufficient to represent both skill blending and grounding in a conversation of finite length~\citep{madotto2020adapter}. Nevertheless, we note that skill blending and grounding are not contradictory, as some skill-grounded utterances leave room for natural shift between skills. Given an utterance ``\textsl{I like sneakers because it is comfortable.}'' which represents skill type P, it seems reasonable to annotate an utterance with skill type K ``\textsl{It is because sneakers were primarily designed for sports.}'' for next dialogue turn. This example further implies that different skills can be blended naturally so that the chatbots learn to provide reasonable responses in a multi-skill dialogue~\citep{roller2020opendomain}.

\section{\textsc{BotsTalk} Framework}

We now present \textsc{BotsTalk}, a novel framework that automatically annotates multi-skill dialogues based on multiple single-skill dialogue datasets. The focus of our framework is to mimic a natural conversation by featuring both skill blending and grounding within a dialogue episode. Figure~\ref{fig:BotsTalk} illustrates three main phases of the framework. Implementation details are provided in Appendix~\ref{sec:implementation}.

\begin{figure*}[t!]
\centering
    \includegraphics[width=130mm]{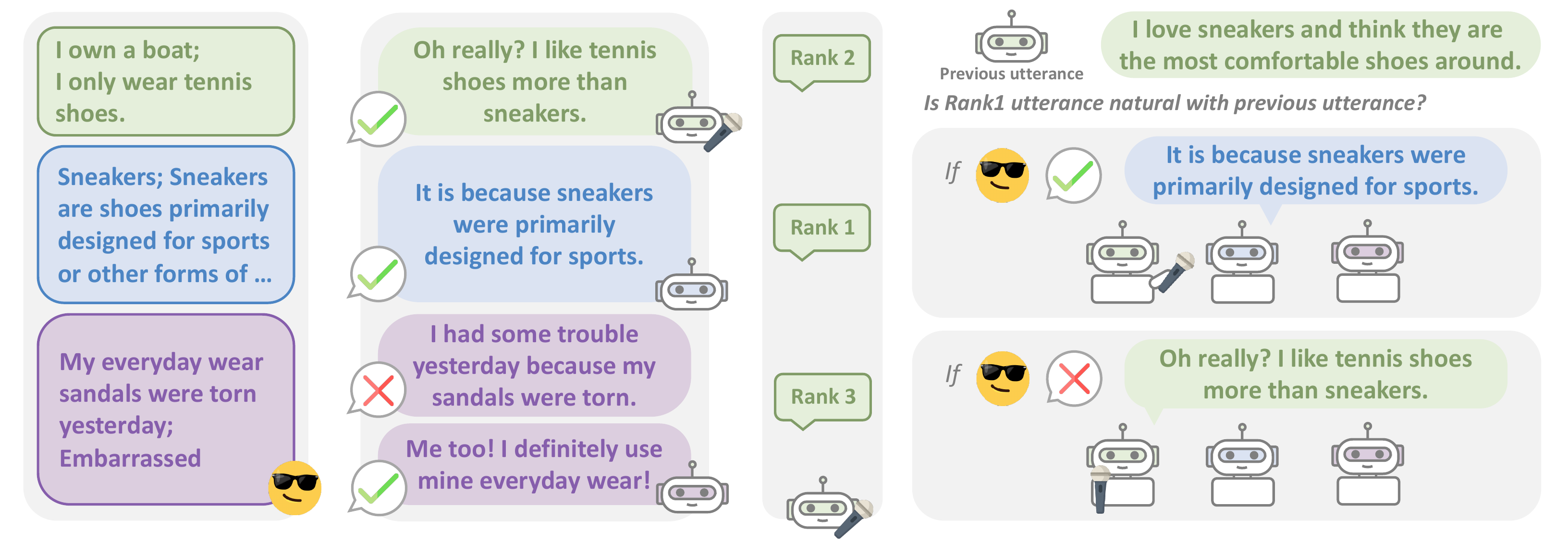}
\caption{Illustration of \textsc{BotsTalk} framework. Green, blue, and purple indicate skill types of P, K, E, respectively.}
\label{fig:BotsTalk}
\end{figure*}

\subsection{Participants in \textsc{BotsTalk}}

In our framework, multiple participants engage in a conversation to iteratively generate desirable multi-skill dialogues.

\vspace{1mm}
\noindent\textbf{Skill Agents \agent} The first participants are multiple single-skill agents who annotate the appropriate skill-grounded utterances to the dialogue. Formally, based on $\mathcal{D}_{m}$ for skill $m$, when given skill context $stx_m$, dialogue context $dtx_t$, and response space $\mathbb{U}$, a skill agent has dialogue models $f:(stx_m,dtx_t) \mapsto \mathbb{U}$ which return a response
\begin{equation}\label{eq:f}\small
    res_{m, t} = f(stx_m,dtx_t;\theta^m)
\end{equation}
where $\theta^m$ is the parameters learned for skill $m$.

We design two main functions of the skill agent, generator model and ranker model, parameterized as $\theta_{gen}^m$ and $\theta_{rnk}^m$ for skill $m$, respectively. For $\theta_{gen}$, we aim to generate responses from response space $\mathbb{U}$ in a token-by-token manner, and thus employ a dodecaDialogue~\citep{shuster2020dodeca} model, a modification of a transformer Seq2Seq architecture. On the other hand, for $\theta_{rnk}$, we consider the response space $\mathbb{U}$ as a list of alternatives to pick the correct response, and thus employ a poly-encoder~\citep{humeau2019poly} model, a transformer-based retrieval architecture, to score and rank response candidates. Both $\theta_{gen}$ and $\theta_{rnk}$ are fine-tuned on individual single-skill datasets\footnote{On the average, generator and ranker models show around 10 perplexity and 90 accuracy on their respective datasets.}.

While all skill agents simulate what response to annotate, only one skill agent is given priority over other skill agents, to ``speak'' the response per dialogue turn for the dialogue annotation, conditioned on a set of skill contexts $\tilde{stx}$ and the dialogue context $dtx_t$. We call this \emph{active agent}. This priority may be passed to another skill agent such that the current active agent is deactivated, and another skill agent will be newly activated to speak.

\vspace{1mm}
\noindent\textbf{Moderator Agent \moderator} A critical constraint for skill agents is that neither the generator nor the ranker for skill $m$ is able to read other skill contexts in $\tilde{stx}$ for different skills. For a skill agent, considering all possible skill contexts in multi-skill dialogues is non-trivial. Instead, as an omniscient oracle for all skill contexts $\tilde{stx}$, we aim to develop another participant named moderator agent, which mediates the conversational flow for desirable multi-skill dialogue annotation. To examine the relevance of the response $res_t$ with all skill contexts $\tilde{stx}$ or the dialogue context $dtx_t$, the moderator agent has a decision function $g:(\tilde{stx},dtx_t,res_t) \mapsto \mathbb{A}$ where  $\mathbb{A}$ is an action space (\ie, approval or refusal) for the given response.

\subsection{Phase 1: Simulate what to speak}

We integrate different dialogue setups from multiple single-skill datasets as seed information to start a conversation (detailed in Appendix~\ref{sec:skill}). For a dialogue episode, dialogue context is initialized as an utterance pair (\ie, two dialogue turns) randomly sampled from a single-skill dataset $\calD_m$, and the skill agent for skill $m$ becomes the initial active agent. Then, for a generalizable dialogue setup, we retrieve the most relevant skill contexts from each of all input datasets $\calD_1, ... ,\calD_M$ for the seed dialogue context with TF-IDF~\citep{chen2017reading}\footnote{While we use a simple IR baseline as lower bound since it is not our main focus, one can easily try different IR system.}.

In the first phase of \textsc{BotsTalk}, all skill agents simulate their own responses for the next dialogue turn. Formally, given a skill context $stx_m$ and the current dialogue context $dtx_t$ in a dialogue episode, a skill agent for skill $m$ generates a plausible response $res_{m,t}$ as
\begin{equation}\label{eq:phase1}\small
    res_{m,t} = \aggregate{argmax}{res_t \in \mathbb{U}} P(res_t|stx_m,dtx_t;\theta^m_{gen}) \cdot g(\tilde{stx}, res_t)
\end{equation}
where $g(\cdot)$ is the function of the moderator agent, which we discuss in the subsequent section.

Depending on individual skills, every skill agent returns its skill-relevant response. For example, as shown in Figure~\ref{fig:BotsTalk}, when ``\textsl{I love sneakers and think they are the most comfortable shoes around.}'' is given as $dtx$, the skill agent for skill P generates a personal response ``\textsl{Oh really? I like tennis shoes more than sneakers.}'' as ${res}_\text{P}$ based on a given persona. Meanwhile, the skill agents for skill K and E generate a knowledgeable response ``\textsl{It is because sneakers were primarily designed for sports.}'' as ${res}_\text{K}$ and a empathetic response ``\textsl{Me too! I definitely use mine everyday wear!}'' as ${res}_\text{E}$.

\subsection{Phase 2: Check dialogue consistency}\label{sec:4.3}

It is well known that neural dialogue systems lack consistency~\citep{li2016persona,welleck2018dialogue}.
Furthermore, as a skill agent uses the specific skill context $stx_m$ instead of $\tilde{stx}$ for response generation, the response is more likely to be semantically in conflict with other skill contexts in $\tilde{stx}$. Suppose a ${stx}_\text{P}$ is ``\textsl{I wear sneakers everyday}'' and a ${res}_\text{E}$ is ``\textsl{I had some trouble yesterday because my sandals were torn}''. This response is inappropriate because ``\textsl{yesterday because my sandals were torn}'' is contradictory to ``\textsl{I wear sneakers everyday}''. Therefore, the moderator agent, who has access to all skill contexts $\tilde{stx}$, filters out conflicting response candidates to preserve dialogue consistency.

Specifically, the moderator agent leverages natural language inference (NLI), a task of determining whether a hypothesis sentence can be inferred from the given premise sentence. The hypothesis sentence is classified into three categories: \textsc{Entail} (true), \textsc{Neutral} (undetermined), and \textsc{Contradict} (false). Based on the NLI classifier, the decision function of the moderator agent is defined as
\begin{equation}\small
    g(\tilde{stx}, res_{t}) = \begin{cases}\small
    1, & \mathrm{NLI}(\tilde{stx}, res_{t}) \not\rightarrow \textsc{Contradict} \\
    0, & \text{otherwise}
    \end{cases}
\end{equation}
which represents approval/refusal of $res_t$ conditioned on $\tilde{stx}$. A skill agent for skill $m$ repeatedly generates new response candidates until its response is approved, as described in Equation~\ref{eq:phase1}.

\begin{figure}[t!]
\centering
    \includegraphics[width=0.7\linewidth]{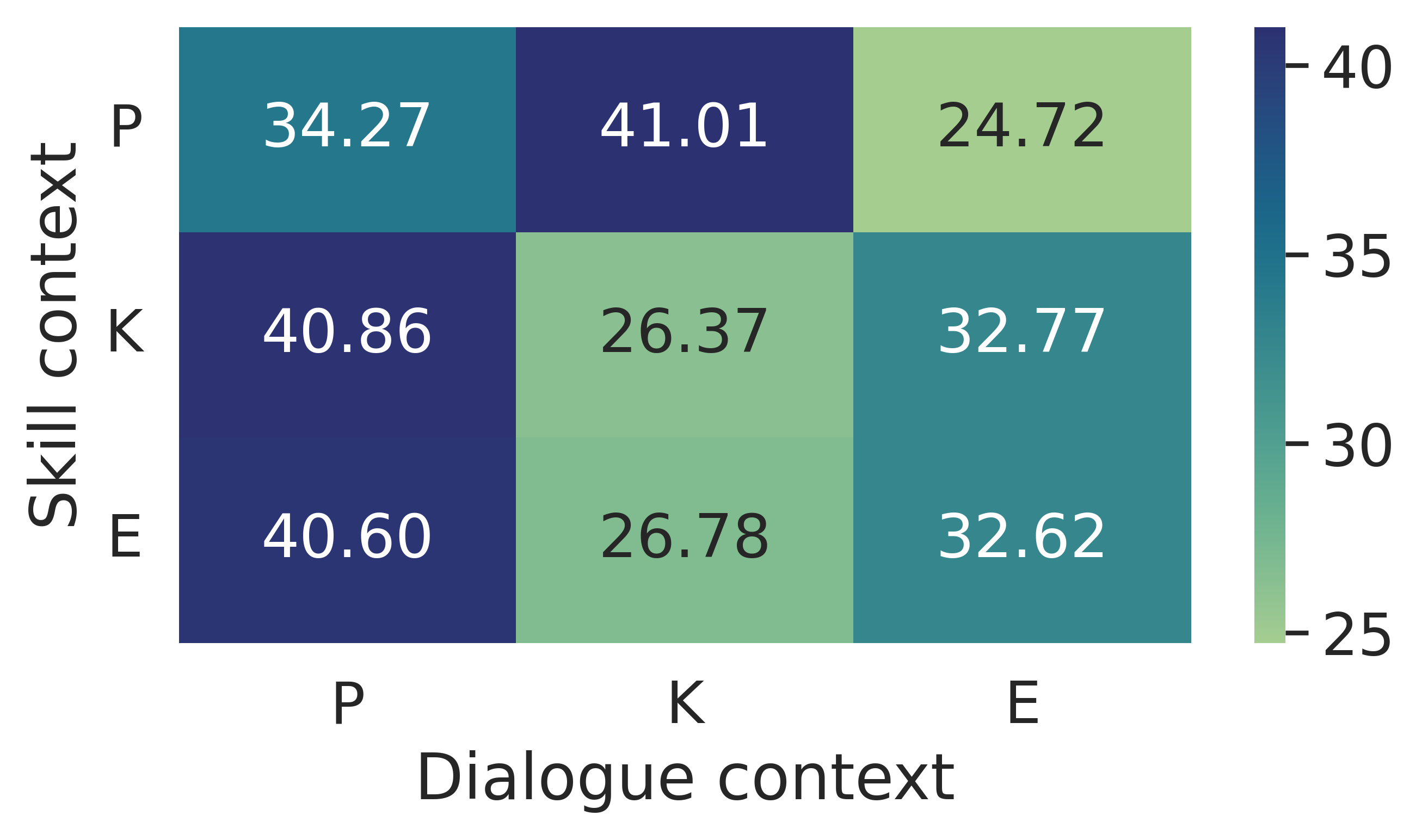}
\caption{Percentages of utterances which are classified as \textsc{Contradict} via NLI classifier, broken down by the type of skill contexts.}
\label{fig:phase2}
\end{figure}

For NLI classifier, we use a RoBERTa~\citep{liu2019roberta} model trained on MNLI~\citep{adina2018mnli}\footnote{Dialogue NLI~\citep{welleck2018dialogue} is biased to ConvAI2.}, which is widely used in fact checking systems~\citep{kim2021robust}\footnote{The RoBERTa model shows 90.59 accuracy on MNLI.}. Overall, about 50\% of utterances are classified as \textsc{Contradict} by NLI classifier. Out of all utterances classified as \textsc{Contradict}, about 70\% are in conflict with other types of skill contexts (Figure~\ref{fig:phase2}). The result demonstrates that skill agents indeed generate inconsistent responses due to the restricted access to other skill contexts. We also find that the overall proportion of utterances conflicting with ${stx}_\text{P}$ is relatively high, apparently because $stx_\text{P}$ contains more distinct descriptions than $stx_\text{K}$ and $stx_\text{E}$.

\subsection{Phase 3: Speak or pass the mic}

The objective of the last phase is to score a set of response candidates and select a final response when given the skill contexts and dialogue context. To this end, we leverage the active agent and the moderator agent, taking into account a balance between skill blending and skill grounding.

\begin{figure}[t!]
     \centering
     \begin{subfigure}[t]{0.22\textwidth}
         \includegraphics[width=\textwidth]{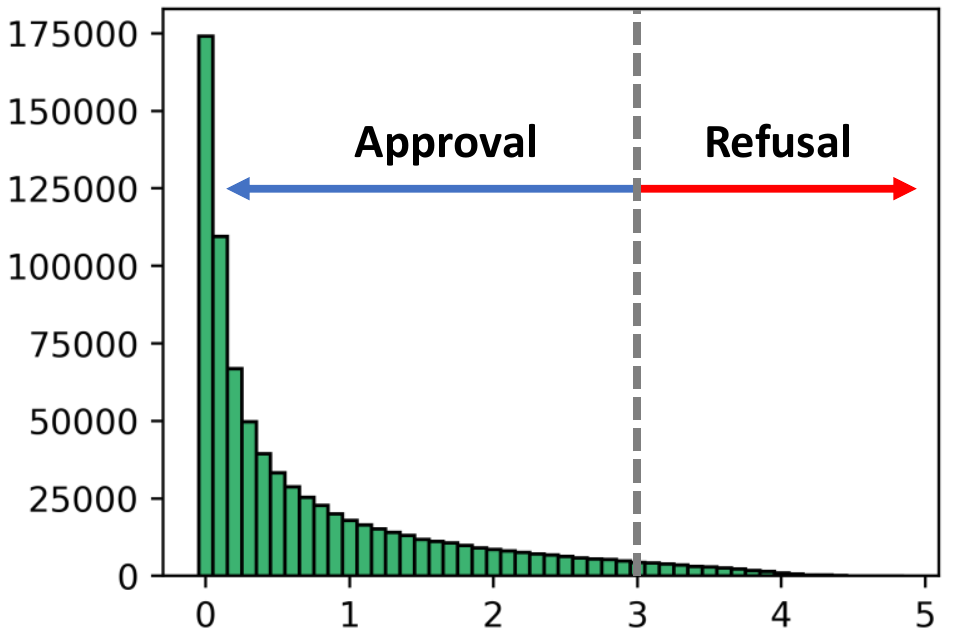}
         \caption{KLD distribution}
         \label{fig:kld}
     \end{subfigure}
     \hfill
     \begin{subfigure}[t]{0.22\textwidth}
         \includegraphics[width=\textwidth]{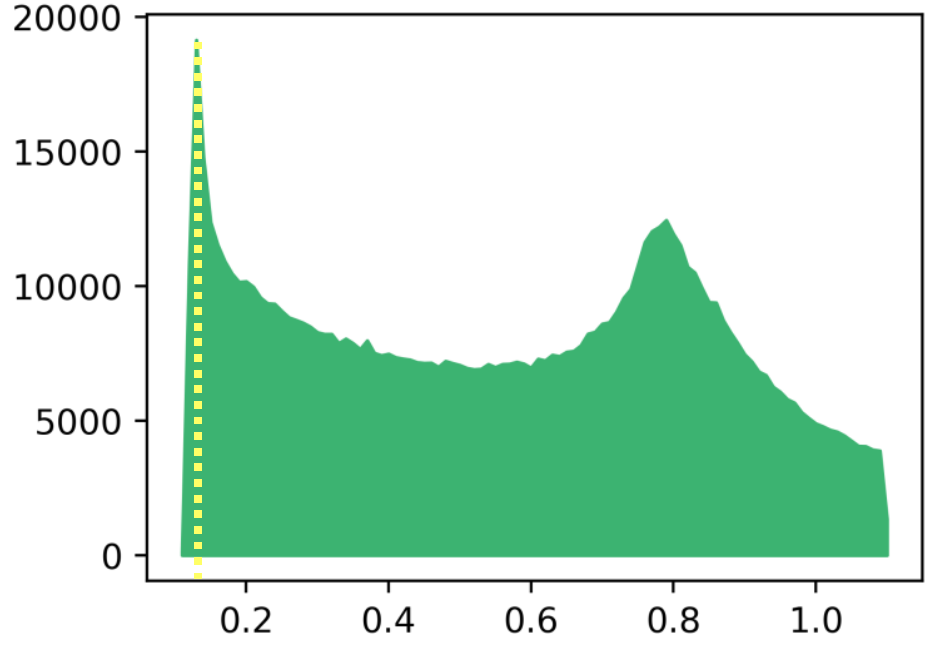}
         \caption{Entropy distribution}
         \label{fig:entropy}
     \end{subfigure}
\caption{KL divergence between skill distributions of consecutive utterances (left) and entropy of skill distributions for all utterances (right).}
\label{fig:phase3}
\end{figure}

\begin{table*}[t!]
\begin{center}\small
{
\begin{tabular}{l | l}
    \toprule
    \textbf{Skill context from ConvAI2} & \textbf{Skill context from ConvAI2} \\
    I like to play soccer; I like to read; ... & I have 3 children; I am a karate black belt; ... \\
    \textbf{Skill context from WoW} & \textbf{Skill context from WoW} \\
    Nike Inc. & Nike Inc.; ... multinational corporation ... Air Jordan ... \\
    \textbf{Skill context from ED} & \textbf{Skill context from ED}  \\
    I really like this girl at my job, but I am ... ; Apprehensive & None \\
    \midrule
    \multicolumn{2}{l}{\textbf{Dialogue}} \\
    \multicolumn{2}{l}{A: Do you have much experience using the different types of cleats? Which do you like best? (P)} \\ 
    \multicolumn{2}{l}{B: I have a little. I also know about Air jordans, a brand of footwear also popular with athletes. (K)} \\
    \multicolumn{2}{l}{A: I enjoy Air jordans as well. I like to play soccer, and it's extremely hard to get good footwear. (P)} \\
    \multicolumn{2}{l}{B: I agree. Air jordans are good for calves, and it's really easy to pull off. (K)} \\ 
    \multicolumn{2}{l}{A: Air jordans are generally made with the most material so it makes sense they'd be easy to use. (K)} \\
    \multicolumn{2}{l}{B: And now shoes can go together with clothes as well, like any other type of material. (K)} \\   
    \multicolumn{2}{l}{A: Do you know much about aeros then? They have excellent fit and beauty. (K)} \\
    \multicolumn{2}{l}{B: I have a hard time finding it but they are great shoes. (P)} \\  
    \multicolumn{2}{l}{A: I hope you can find ones that are comfortable to you. (E)} \\
    \multicolumn{2}{l}{B: Yes. I hope you can get those shoes too. (E)} \\
    \bottomrule
\end{tabular}
}
\end{center}
\caption{A conversation from the BS$\mathbb{B}$T dataset. Speaker A is given five personas, one topic, and a situation with an emotion (top left), while speaker B is given five personas, one topic, and seven knowledge resources (top right).}
\label{tab:bsbt_dialogue_example}
\end{table*}

Let $\mathbb{U}_{res}$ be the set of response candidates $res_{1,t},...,res_{M,t}$ from all skill agents. The active skill agent identifies the most appropriate response $res_t^*$ in $\mathbb{U}_{res}$ based on its ranker model $\theta^m_{rnk}$, then asks the moderator agent to attach the selected response into the next dialogue context $dtx_{t+1}$ for annotation. Formally, we define such process as
\begin{equation}\label{eq:active}\small
    res_t^* = \aggregate{argmax}{res_t \in \mathbb{U}_{res}}  P(res_t | stx_m, dtx_t; \theta^m_{rnk}) \cdot g(dtx_t, res_t)\\
\end{equation}
where $g(\cdot)$ is the function of the moderator agent. To compute $g(dtx_t, res_t)$, the moderator agent adopts a skill classifier $\calP$ that identifies corresponding skill for the response. We use a BERT~\citep{devlin2018bert} model trained on utterances in $\calD_m$ and their corresponding skill labels $m$ for all skill types $\mathbb{M}$\footnote{The BERT model shows 81.95 accuracy at inference time.}. Once $\calP$ is learned, the decision function of the moderator agent is defined as
\begin{equation}\small
    g(dtx_t, res_t) = \begin{cases}\small
    1,& \mathrm{KL}(\calP(res_{t-1}^*) || \calP(res_t)) < \alpha \\
    0,& \text{otherwise}
    \end{cases}
\end{equation}
where $res_{t-1}^*$ is the last utterance of $dtx_t$ and $\calP(\cdot) \in \mathbb{R}^M$ outputs the skill distribution of the response. Based on KL divergence between two distributions, $g(dtx_t, res_t)$ is discretized as the approval/refusal decision by a pre-defined threshold $\alpha$ (Figure~\ref{fig:kld}). Once the moderator agent accepts the candidate $res_t$ from an inactive agent as the final response, the active agent passes the mic, or the priority for annotation, to the inactive agent.

In practice, we compute entropy of the skill distributions of all utterances to investigate whether there is room for shifting between skills. The value of entropy indicates the uncertainty of the skill type of an utterance: utterances with high entropy values are uncertain, generic responses. Figure~\ref{fig:entropy} shows that the number of generic utterances is far from negligible, suggesting that there are opportunities to shift to other skills and thus both skill blending and grounding can be satisfied in a conversation.

\section{Blended Skill BotsTalk (BS$\mathbb{B}$T)}

\subsection{Data Statistics}

\begin{table}[t!]
\begin{center}\small
{\begin{tabular}{l r r r r}
    \toprule
    \textbf{Dataset} & \textbf{Dialogues} & \textbf{Utterances} & \textbf{Turns}\\
    \midrule
    ConvAI2  & 19,893 & 145,873 & 7 \\
    WoW      & 22,311 & 201,999 & 9 \\
    ED       & 24,850 & 51,245 & 2 \\
    BST      & 6,808 & 58,575 & 8 \\
    BS$\mathbb{B}$T & 300,000 & 3,000,000 & 10 \\


    \bottomrule
\end{tabular}}
\end{center}
\caption{Statistics of dialogue datasets: the number of dialogues, utterances, and the average number of turns.}
\label{tab:dialogue_dataset_stat}
\end{table}

We collect a multi-skill dialogue dataset, namely Blended Skill BotsTalk (BS$\mathbb{B}$T), using \textsc{BotsTalk}. The dataset consists of 300K dialogues with 3M utterances. Each utterance is labeled using the skill classifier with skill annotation (\eg, personality from ConvAI2, knowledge from WoW, or empathy from ED), including skill type and skill distribution. Table~\ref{tab:bsbt_dialogue_example} presents an example from BS$\mathbb{B}$T. As shown in Table~\ref{tab:dialogue_dataset_stat}, one of the salient features of BS$\mathbb{B}$T is its scalability, mainly because BS$\mathbb{B}$T is composed of bot-bot conversations collected through a machine-sourced approach while other datasets comprise crowdsourced human-to-human conversations.

\begin{figure}[t!]
     \centering
     \begin{subfigure}[t]{0.24\textwidth}
         \includegraphics[width=\textwidth]{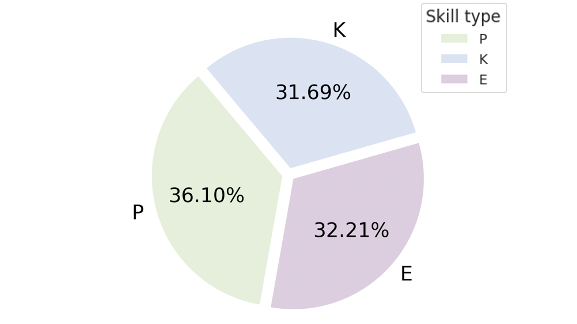}
         \caption{Utterances}
         \label{fig:skill_percentage}
     \end{subfigure}
     \hfill
     \begin{subfigure}[t]{0.22\textwidth}
         \includegraphics[width=\textwidth]{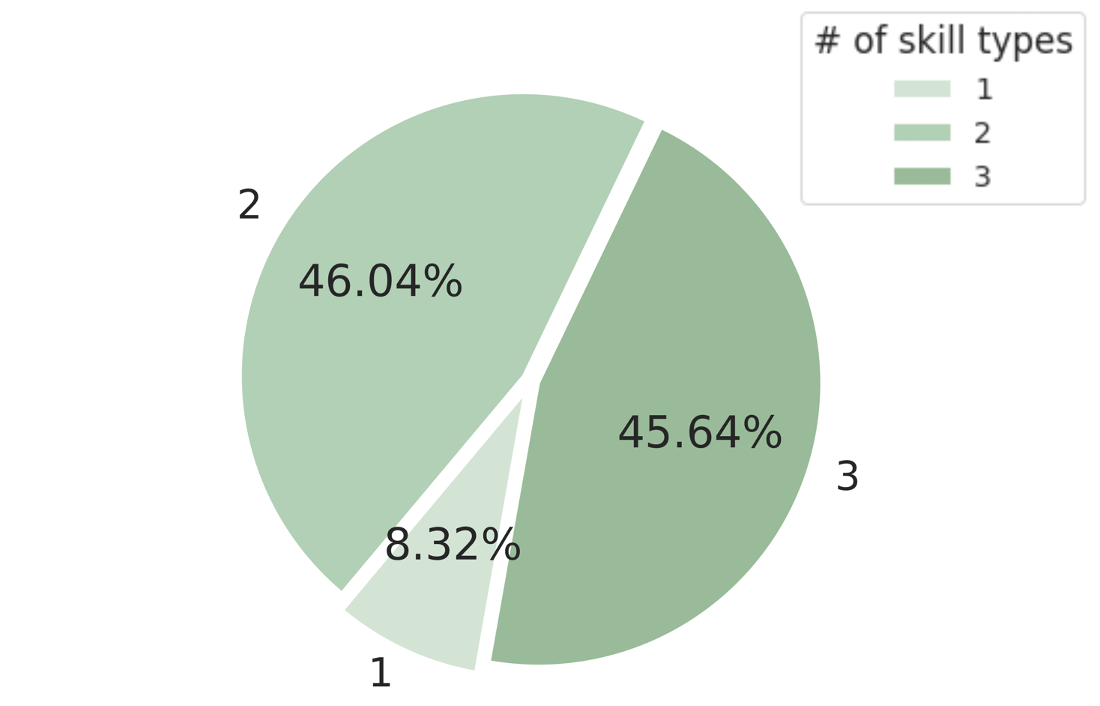}
         \caption{Dialogues}
         \label{fig:skill_count}
     \end{subfigure}
\caption{Percentages of utterances with respect to the skill type (left) and dialogues broken down by the number of skill types in the dialogue (right).}
\label{fig:skill_blending}
\end{figure}


\vspace{1mm}
\noindent\textbf{Skill Blending} Figure~\ref{fig:skill_blending} summarizes the results of skill annotation in BS$\mathbb{B}$T dataset. Overall, the skill annotation percentages are 36.10\% for personality, 31.69\% for knowledge, and 32.21\% for empathy (Figure~\ref{fig:skill_percentage}). Moreover, over 90\% of the dialogues demonstrate at least 2 of the 3 skills within a single dialogue (Figure~\ref{fig:skill_count}), suggesting the vast majority of conversations feature more than one skill type.

\begin{figure}[t!]
\centering
    \includegraphics[width=0.7\linewidth]{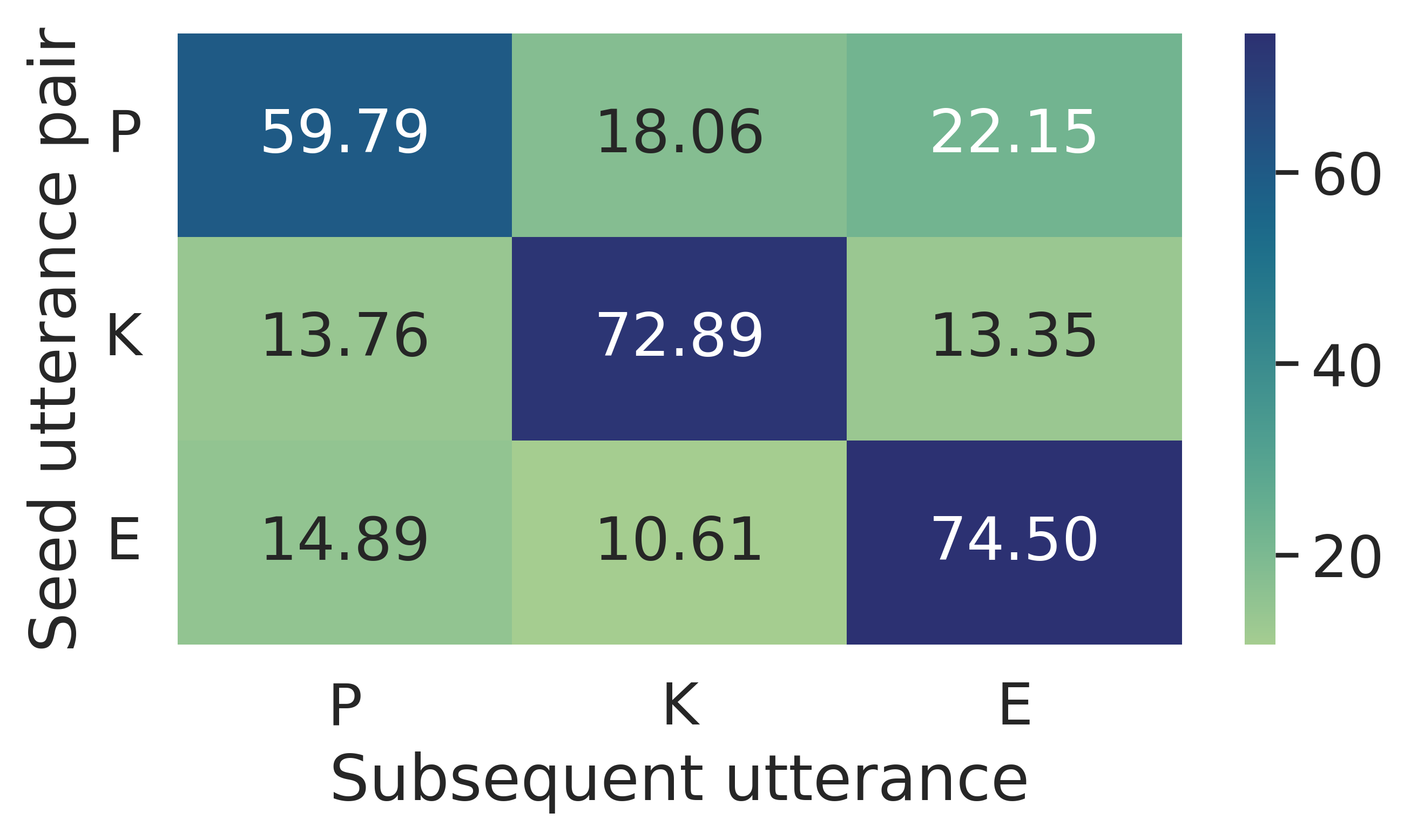}
\caption{Percentages of the skill type of utterances subsequent to the utterance pairs, broken down by provenance skills of the utterance pairs.}
\label{fig:skill_bias_seed_utterance}
\end{figure}

\noindent\textbf{Skill Grounding} Although we focus on blending skills, multi-skill dialogue datasets should also contain sufficient sessions grounded to specific skills in conversations for a model to learn the ability of skill grounding. For that, we explore the continuity of skills by investigating skill types of utterances subsequent to seed utterance pairs whose provenance skills are determined by their original datasets. For all skills, more than half of the utterances followed by the utterance pairs are labeled as the same skill types of the utterance pairs (Figure~\ref{fig:skill_bias_seed_utterance}).

\subsection{Experimental Setups}

We conduct a set of experiments to test our BS$\mathbb{B}$T over BST benchmark through automatic and human evaluation. To the best of our knowledge, BST benchmark is the only multi-skill dialogue benchmark which gauges how successful a model is at blended objective as well as grounded objective.

Following~\citet{smith2020bst}, we consider the retrieval task as our primary task and adopt a 256-million parameter poly-encoder~\citep{humeau2019poly} pre-trained on pushshift.io Reddit dataset as a base architecture. We further include the generative task as our secondary task and adopt a pre-trained BART~\citep{lewis2020bart} as a base architecture. We fine-tune these base architectures on individual datasets, \ie, ConvAI2, WoW, ED, BST, and BS$\mathbb{B}$T, and use them as our baselines. We describe implementation details in Appendix~\ref{sec:baselines}.

For the retrieval task, we report recall@k (R@k), where each test example has 100 possible candidates to select from, as well as mean reciprocal rank (MRR). For the generative task, we compute the average score of BLEU-1, -2, -3, -4 (Avg. BLEU).

\begin{table}[t!]
\begin{center}\small
{
\begin{tabular}{l ccc c}
    \toprule
    & \multicolumn{3}{c}{Retrieval} & \multicolumn{1}{c}{Generative} \\
    \cmidrule(lr){2-4} \cmidrule(lr){5-5}
    Model & R@1 & R@5 & MRR & Avg. BLEU \\
    \midrule
    ConvAI2 & 75.92 & 94.04 & 83.96 & 3.75 \\
    WoW & 67.48 & 89.57 & 77.11 & 4.08  \\
    ED & 65.96 & 88.69 & 76.10 & 3.15 \\
    BST & 75.92 & 94.76 & 84.14 & 4.31 \\
    BS$\mathbb{B}$T & \textbf{80.68} & \textbf{95.79} & \textbf{87.39} & \textbf{4.38} \\
    \bottomrule
\end{tabular}
}
\end{center}
\caption{Automatic evaluation on BST benchmark.}
\label{tab:evaluation_blending}
\end{table}

\subsection{Automatic Evaluation}

The results of retrieval and generative models on BST benchmark are shown in Table~\ref{tab:evaluation_blending} (detailed results are in Appendix~\ref{sec:additional}). We observe that multi-skill models, \ie, BST and BS$\mathbb{B}$T models, are superior to single-skill models, \ie, ConvAI2, WoW, and ED models on the BST benchmark. As multi-skill dialogues require an understanding of both skill blending and grounding, single-skill models who are only grounded to each of skills struggle to seamlessly blend them over the course of a conversation, whereas multi-skill models are able to not only exhibit individual skills but also combine different skills in a conversational flow. In particular, BS$\mathbb{B}$T model outperforms all of the baselines on all automatic metrics. This indicates that our machine-sourced dataset works properly as the training resource to learn the ability of blending skills as well as grounding to various skills.

\begin{figure}[t!]
\centering
    \includegraphics[width=\linewidth]{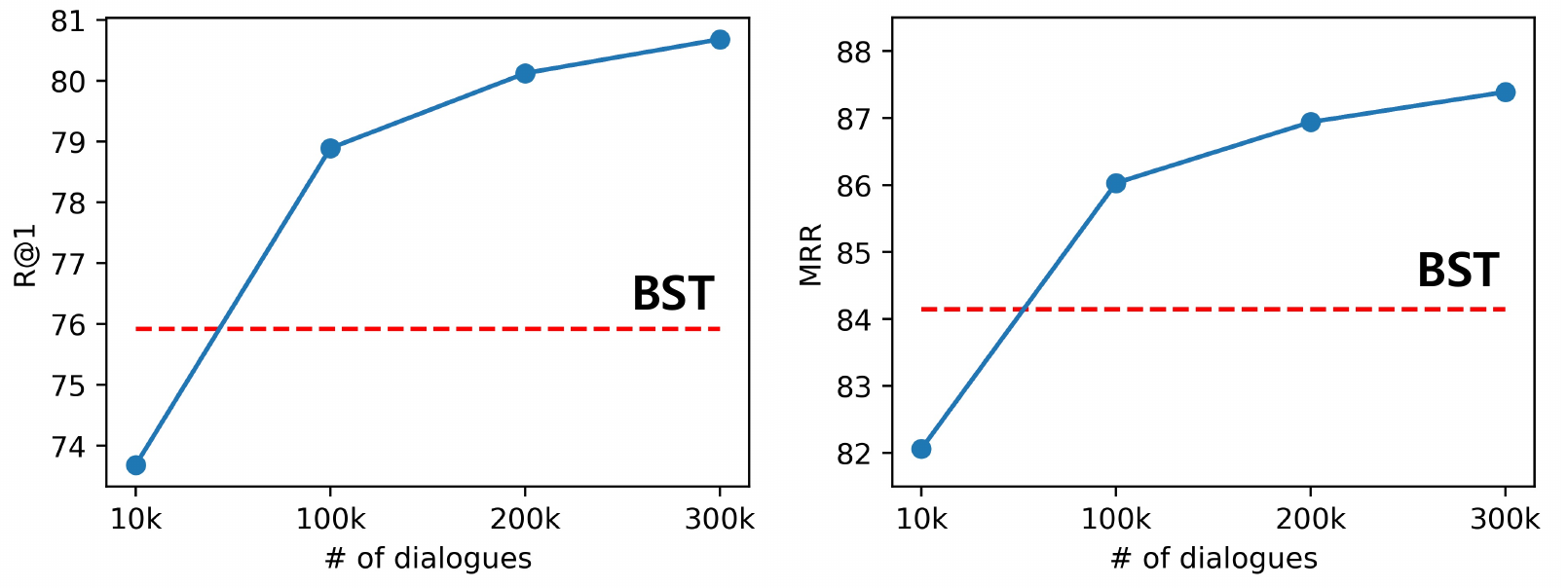}
\caption{The effect on model performance by varying the number of dialogues, measured by R@1 and MRR.}
\label{fig:performance_bsbt_size}
\end{figure}



To explore the impact of BS$\mathbb{B}$T size on the model performance, we fine-tune the retrieval  architecture on the BS$\mathbb{B}$T datasets of varying scales. Figure~\ref{fig:performance_bsbt_size} illustrates the performance of BS$\mathbb{B}$T model in terms of R@1 and MRR when the size of the dataset gradually increases. BS$\mathbb{B}$T300K model achieves a significant performance boost from BS$\mathbb{B}$T10K model, surpassing BST model and showing the best performance. This result not only affirms the importance of large-scale training for building multi-skill chatbots but also indicates the potential of BS$\mathbb{B}$T dataset, as our dataset is collected by automatic \textsc{BotsTalk} framework without human intervention (\ie, no manual annotation or verification).

\begin{table}[t!]\small\centering
\begin{tabular}{l|ccc}
    & BST & vs. & BS$\mathbb{B}$T \\
    & (Win \%) & & (Win \%) \\
    \midrule
    Engagingness & \colorbox{lightblue}{43} && \colorbox{deepblue}{\textcolor{white}{57}} \\
    Interestingness & \colorbox{lightblue}{47} && \colorbox{deepblue}{\textcolor{white}{53}} \\
    Humanness & \colorbox{lightblue}{44} && \colorbox{deepblue}{\textcolor{white}{56}} \\ 
\end{tabular}
\caption{Human evaluation for pairwise comparison between BST and BS$\mathbb{B}$T datasets.}
\label{tab:data_quality}
\end{table}

\subsection{Human Evaluation}\label{sec:human_evaluation}


To assess the quality of BS$\mathbb{B}$T dataset, we perform human evaluation by employing ACUTE-Eval~\citep{li2019acute}, a popular metric for multi-turn dialogue evaluation~\citep{dinan2020queens,li2020don}. We randomly sample 100 dialogues from the BST and BS$\mathbb{B}$T datasets respectively, and then ask human evaluators to compare each pair of dialogues over three axes: engagingness, interestingness and humanness. We provide more details for the specific settings in Appendix~\ref{sec:human}. 
Table~\ref{tab:data_quality} shows that for all metrics, BS$\mathbb{B}$T dataset achieves comparable and even slightly higher win percentages over BST dataset. This ensures the quality of BS$\mathbb{B}$T and thus validates that \textsc{BotsTalk} framework can be a viable alternative to crowdsourcing when constructing multi-skill dialogue datasets.

\noindent\textbf{Qualitative Analysis} 
Although most of conversations from BS$\mathbb{B}$T satisfies the desirable characteristics for multi-skill dialogues as evidenced by a set of experiments, any side effect may occur since the dialogues are collected through automatic annotation. Therefore, we select best-case and worst-case examples (each in Table~\ref{tab:bsbt_cherry} and Table~\ref{tab:bsbt_lemon}) and provide empirical results including three types of error cases. First, in a few cases, speakers repeat greeting (about 3 turns) at the end of conversation. This is mainly because we set dialogues to be of fixed length, while the conversation may end earlier than the given turns. Second, as observed in Figure~\ref{fig:skill_count}, we find a number of dialogues that only features one skill type. Lastly, some responses tend to show little relevance with the skill contexts of their corresponding skills, although they are grammatically sound and meaningfully move the conversations forward. We give a deeper analysis in Appendix~\ref{sec:qualitative}.

\begin{table}[t!]
\begin{center}\small
{
\begin{tabular}{l ccc}
    \toprule
    & \multicolumn{3}{c}{Retrieval} \\
    \cmidrule(lr){2-4} 
    Model & R@1 & R@5 & MRR \\
    \midrule
    MTL & 78.95 & 95.37 & 86.23 \\
    BS$\mathbb{B}$T & 80.68 & 95.79 & 87.39 \\
    MTL + BS$\mathbb{B}$T100K & 80.94 & 95.79 & 86.92 \\
    MTL + BS$\mathbb{B}$T200K & 82.01 & 96.37 & 87.83 \\
    MTL + BS$\mathbb{B}$T300K & 82.10 & 96.79 & 88.04 \\
    
    \bottomrule
\end{tabular}
}
\end{center}
\caption{Performance of MTL models on BST benchmark, reported by R@1, R@5, and MRR.}
\label{tab:evaluation_mtl}
\end{table}

\subsection{Analysis on Multi-task Learning}


Given an access to multiple single-skill dialogue datasets, a straightforward approach of developing a multi-skill chatbot is to multi-task on all of them during fine-tuning step~\citep{shuster2020dodeca,roller2021recipes}. Therefore, we consider MTL model, a poly-encoder~\citep{humeau2019poly} pre-trained on pushshift.io Reddit and fine-tuned in multi-task fashion across ConvAI2, WoW, and ED. We further fine-tune the MTL model on BS$\mathbb{B}$T datasets of varying scales sequentially, to probe the effectiveness of BS$\mathbb{B}$T as a training resource for multi-task training scheme. In Table~\ref{tab:evaluation_mtl}, MTL model lags behind BS$\mathbb{B}$T model on BST benchmark, but performs noticeably better when fine-tuned on BS$\mathbb{B}$T dataset in addition. Such improvement in the performance is an encouraging sign that BS$\mathbb{B}$T is orthogonally applicable to multi-tasking strategy. We observe that the performance gain becomes marginal when the size of the dataset increases. We hypothesize that as multi-task learning and BS$\mathbb{B}$T are parameterized and materialized knowledge for multi-skill dialogues respectively, there can be an overlap between the knowledge dialogue systems learn. We leave the mitigation of such overlap for future work. Nevertheless, MTL model achieves the best performance when fine-tuned on BS$\mathbb{B}$T300K dataset, as it compensates the overlap with its size.

\begin{table}[t!]
\begin{center}\small
{
\begin{tabular}{l ccc}
    \toprule
    Model & ConvAI2 & WoW & ED \\
    \midrule
    \multicolumn{4}{l}{\textbf{Retrieval}} \\
    ConvAI2 & 0 & $-$10.95 & $-$14.91 \\ 
    WoW & $-$30.56 & 0 & $-$16.95 \\
    ED  & $-$27.15 & $-$10.74 & 0 \\
    BST & $-$14.33 & $-$8.67 & $-$14.70 \\
    BS$\mathbb{B}$T & \textbf{$-$2.12} & \textbf{$-$1.92} & \textbf{$-$1.52} \\
    \midrule
    \multicolumn{4}{l}{\textbf{Generative}} \\
    ConvAI2 & 0 & $-$3.27 & $-$2.89 \\
    WoW & $-$3.60 & 0 & $-$2.51 \\
    ED  & $-$3.74 & $-$3.12 & 0 \\
    BST & $-$2.68 & $-$2.78 & \textbf{$-$1.86} \\
    BS$\mathbb{B}$T & \textbf{$-$0.98} & \textbf{$-$2.19} & $-$1.95 \\
    \bottomrule
\end{tabular}
}
\end{center}
\caption{Performance on single-skill benchmarks, measured by $\Delta_\text{R@1}$ for retrieval models and $\Delta_\text{Avg. BLEU}$ for generative models.}
\label{tab:evaluation_single}
\end{table}

\subsection{Analysis on Skill Grounding Ability}\label{sec:evaluation_single}


To gain more insights into individual skill grounding ability, we evaluate the baselines on single-skill benchmarks, \ie, ConvAI2, WoW, and ED benchmarks (detailed results are in Appendix~\ref{sec:additional}). We compute a relative performance drop $\Delta_\text{R@1}$ for retrieval models and $\Delta_\text{Avg. BLEU}$ for generative models over the best performing model on the respective benchmark, which gives us upper bound we aim for with our model. Results are summarized in Table~\ref{tab:evaluation_single}. As expected, each of single-skill models perform best on their original benchmarks but not as well on other benchmarks, whereas the multi-skill models show more well-rounded performance across all benchmarks. In particular, BS$\mathbb{B}$T model outperforms BST model on most cases and even achieves comparable performance to single-skill models on corresponding single-skill benchmarks. This suggests that BS$\mathbb{B}$T is effective to not only inject the ability of blending various skills but also maintain the ability for grounding specific skill.

\section{Conclusion}

To build multi-skill chatbots, we construct a large-scale dialogue dataset BS$\mathbb{B}$T through automatic \textsc{BotsTalk} framework. We validate its efficacy as training resource by experiments and analyses.

\section{Limitations}

We summarize the error patterns in BS$\mathbb{B}$T and discuss potential directions to improve \textsc{BotsTalk} framework. First, our framework \emph{always} produces a fixed-length conversation, even when the conversation ends earlier than the given turns. As shown in Example 4 from Table~\ref{tab:bsbt_lemon}, this often results in generic and repetitive responses at the end of a dialogue. In future work, we are interested in training dialogue models that understand \emph{when} to end a conversation based on the context. Second, as observed in Figure~\ref{fig:skill_count}, a few dialogues fail to cover explicit transitions between multiple skills. For instance, Example 5 from Table~\ref{tab:bsbt_lemon} only features one skill type and lacks the nature of skill blending, which may hinder models from learning diverse communicative skills. Third, some responses, particularly those near the end of conversations, tend to show little relevance with their corresponding skill contexts. We conjecture that as a conversation flows, skill agents condition their responses more on the dialogue context, which is likely to be longer than the pre-defined skill contexts. To enhance the model's ability of skill grounding, one can employ conditional training by modeling the progress of the conversation and controlling input contexts. We leave these issues for future research.

\section*{Acknowledgements}\label{sec:acknowledgements}
We would like to thank anonymous reviewers for their valuable comments. This research was partially supported by the Institute of Information \& communications Technology Planning \& Evaluation (IITP) grant funded by the Korea government (MSIT) (No. 2020-0-01361, Artificial Intelligence Graduate School Program (Yonsei University)) and the National Research Foundation of Korea (NRF) grant funded by the Korea government (MSIT) (No. 2022-11-0941).

\bibliography{anthology,custom}
\bibliographystyle{acl_natbib}

\appendix

\section{Overview}

In the following sections, we explore more details on \textsc{BotsTalk} framework and BS$\mathbb{B}$T dataset. In Appendix~\ref{sec:single}, we lay out the details of single-skill dialogue datasets and how they are incorporated into \textsc{BotsTalk} framework to construct BS$\mathbb{B}$T dataset. We provide implementation details in Appendix~\ref{sec:implementation} for all component models of participants in \textsc{BotsTalk} framework. We also provide implementation details of baselines used for experiments in Appendix~\ref{sec:baselines}. The evaluation results on all benchmarks used in this paper are in Appendix~\ref{sec:additional}. The specific settings for human evaluation are in Appendix~\ref{sec:human}. Finally, we present a number of conversation examples from BS$\mathbb{B}$T and further analyze its strengths and weaknesses in Appendix~\ref{sec:qualitative}.

\section{Single-skill Datasets into \textsc{BotsTalk}}\label{sec:single}

We describe details on the single-skill dialogue datasets used to construct BS$\mathbb{B}$T and elaborate on how they are incorporated into \textsc{BotsTalk} framework to construct our dataset. Example dialogues from the single-skill dialogue datasets \ie, ConvAI2, WoW, ED, are shown in Table~\ref{tab:conv_dialogue_example}, \ref{tab:wow_dialogue_example}, \ref{tab:ed_dialogue_example}.

To integrate different dialogue setups from the single-skill dialogue datasets, we follow the basic settings for constructing a dialogue dataset, assuming a multi-turn, one-to-one conversation between two speakers. We simulate turn-taking in a conversation by switching two different sets of skill contexts for the input skill context $stx$ to a dialogue model $f$ in a skill agent.

\subsection{ConvAI2}

ConvAI2~\citep{dinan2019convai2} is a dataset based on PersonaChat~\citep{zhang2018persona}. ConvAI2 dataset contains of more than 140K utterances from conversations in which each of paired crowdworkers is given a role based on their persona description and gets to know their partner. Specifically, the speaker pairs are each assigned profiles from a set of 1155 possible personas, each consisting of at least 5 profile sentences. The personas are collected through crowdsourcing, where the workers are asked to create natural, descriptive profiles that contain typical topics of human interest. Workers are also asked to keep each profile sentence short, \ie, no longer than 15 words.

Following the setting from ConvAI2, we define a skill context $stx_\text{P}$ as a profile comprising 5 distinct persona sentences. We then provide two different skill contexts as the input to the dialogue model $f$ in an alternating manner to simulate turn-taking.

\subsection{Wizard of Wikipedia}

Wizard of Wikipedia~\citep{dinan2019wizard} task involves discussing a given topic in depth, where the goal is to both engage the partner as well as display expert knowledge. The dataset consists of 194K utterances over 1250 topics, where each conversation begins with a randomly chosen topic. A retrieval system over Wikipedia is used to retrieve articles from which the dialogues are grounded during the human-human crowdsourced conversations. The topics are also crowdsourced and range from commuting to Gouda Cheese to Arnold Schwarzenegger. Each conversation in the dataset involves two speakers named the apprentice and the wizard: the apprentice aims at delving deeply into a topic whereas the wizard uses knowledge in articles retrieved from Wikipedia to craft a relevant reply. Specifically, given a topic derived from the dialogue context, the apprentice keeps the conversation engaging and talks eagerly about a topic, while the wizard responds to the apprentice based on the first paragraphs of 7 relevant Wikipedia articles provided by the retrieval system. 

In our setting, we use a simpler version of Wizard of Wikipedia task, which ignores the retrieval aspect of the task. We first specify the topic of the conversation, which is the same for the apprentice and wizard. The skill context $stx_\text{K}$ of the apprentice is thus defined as the given topic, while the skill context $stx_\text{K}$ of the wizard is defined as a topic and 7 relevant knowledge sources.

\subsection{Empathetic Dialogues}

Empathetic Dialogues~\citep{rashkin2019ed} is a dataset includes 50K utterances of crowdworker conversations grounded in an emotional situation. In the conversation, one speaker describes a personal situation based on an emotion label and the other speaker, named listener, displays empathy in their response. Specifically, a pair of workers (\ie, speaker and listener) are asked to choose an emotional word each, depict a situation in 1-3 sentences based on the label, and engage in a short conversation of 4-8 utterances about each of the situations. Neither of the workers, whether they be the speaker or the listener, can see the emotion label and the situation description of their partner, so that they must refer only to cues within the conversation for their response.

In our setting, we define the situation description and its corresponding emotion label as the skill context $stx_\text{E}$ of the speaker. Note that we do not define the skill context $stx_\text{E}$ of the listener for our framework, so that the dialogue system is trained to show empathy based solely on the conversation.

\section{Implementation Details of \textsc{BotsTalk}}\label{sec:implementation}

For the implementation of \textsc{BotsTalk} framework, we employ ParlAI\footnote{\url{https://github.com/facebookresearch/ParlAI}} toolkit, which is specialized in training and evaluating dialogue systems. We will release our agents and dataset for public use.

\subsection{Skill Agents}

In our \textsc{BotsTalk} framework, a skill agent leverages both generator model and ranker model.

Given a $stx_k$ and $dtx$ as input, a generator model of skill agent generates a response for the next dialogue utterance. For the generator model, we employ a dodecaDialogue~\citep{shuster2020dodeca}. The dodecaDialogue model is a modification of transformer seq2seq architecture, which has a 8-layer encoder, 8-layer decoder with 512 dimensional embeddings and 16 attention heads. We fine-tune the dodecaDialogue models on ConvAI2, WoW, and ED, respectively. We use nucleus sampling as the decoding strategy for generative models at inference time. The deodecaDialogue model shows 11.19, 8.46, and 11.08 perplexity on ConvAI2, WoW, and ED.

Given a $stx_k$ and $dtx$ as input, a ranker model of skill agent selects the next dialogue utterance by scoring a large set of candidate responses and outputting the one with the highest score. For the ranker model, we employ the poly-encoder architecture of \citet{humeau2019poly}. The poly-encoder encodes global features of the context using multiple representations, which are attended to by each possible candidate response. This final attention mechanism gives improved performance over a single global vector representation whilst still being tractable to compute compared to simply concatenating input and output as input to a Transformer. The poly-encoder has state-of-the-art performance on a number of dialogue tasks when compared to other retrieval models, and also gives comparable performance to the winning generative models on the ConvAI2 competition task in terms of human evaluation.

More specifically, we consider a 256M parameter poly-encoder model, which has 12 layers, 12 attention heads, and a hidden size of 768. We pre-train our poly-encoder on pushshift.io Reddit dataset and then fine-tune on ConvAI2, WoW, and ED, respectively. We use a large number of negatives by considering the other batch elements as negative training samples, avoiding recomputation of their embeddings. We use the Adamax optimizer without weight decay, a learning rate of 5e-5 with batch size 128, epoch 8. The learning rate decays by a factor of 0.4 upon plateau of the loss evaluated on the valid set every half epoch. The best parameters are chosen based on R@1 score. The poly-encoder model shows 89.41, 91.01, and 63.26 R@1 on ConvAI2, WoW, and ED, respectively.

\begin{table*}[t!]
\begin{center}\small
{
\begin{tabular}{l cccc cccc}
    \toprule
    & \multicolumn{4}{c}{Retrieval} & \multicolumn{4}{c}{Generative} \\
    \cmidrule(lr){2-5} \cmidrule(lr){6-9}
    Model & R@1 & R@5 & R@10 & MRR & BLEU-1 & BLEU-2 & BLEU-3 & BLEU-4 \\
    \midrule
    
    \textbf{Evaluation on BST benchmark} \\
    ConvAI2~\citep{zhang2018persona} & 75.92 & 94.04 & 97.19 & 83.96 & 10.97 & 2.88 & 0.86 & 0.32 \\
    WoW~\citep{dinan2019wizard} & 67.48 & 89.57 & 94.33 & 77.11 & 12.00 & 3.20 & 0.80 & 0.31\\
    ED~\citep{rashkin2019ed} & 65.96 & 88.69 & 93.80 & 76.10 & 9.36 & 2.47 & 0.61 & 1.72 \\ 
    BST~\citep{smith2020bst} & 75.92 & 94.76 & 97.83 & 84.14 & 12.19 & 3.65 & 1.06 & 0.37 \\
    BS$\mathbb{B}$T (Ours) & 80.68 & 95.79 & 98.16 & 87.39 & 11.92 & 3.74 & 1.28 & 0.57  \\ 
    \midrule
    
    \textbf{Evaluation on ConvAI2 benchmark} \\
    ConvAI2~\citep{zhang2018persona} & 88.46 & 98.92 & 99.71 & 93.03 & 17.69 & 7.21 & 2.96 & 1.15 \\
    WoW~\citep{dinan2019wizard} & 57.90 & 86.85 & 95.80 & 70.59 & 10.97	& 2.82 & 0.64 & 0.18 \\
    ED~\citep{rashkin2019ed} & 61.31 & 89.44 & 96.69 & 73.53 & 10.64 & 2.60 & 0.63 & 0.20 \\ 
    BST~\citep{smith2020bst} & 74.13 & 95.64 & 98.80 & 83.37 & 12.42 & 4.00 & 1.40 & 0.48 \\
    BS$\mathbb{B}$T (Ours) & 86.34 & 98.00 & 99.42 & 91.46 & 16.12 & 5.88 & 2.12 & 0.81 \\
    \midrule
    
    \textbf{Evaluation on WoW benchmark} \\
    ConvAI2~\citep{zhang2018persona} & 79.84 & 96.97 & 98.84 & 87.62 & 7.88 & 1.89 & 0.53 & 0.15 \\
    WoW~\citep{dinan2019wizard} & 90.79 & 99.28 & 99.66 & 94.67 & 14.85	& 5.63 & 2.02 & 1.02 \\
    ED~\citep{rashkin2019ed} & 80.05 & 96.25 & 98.37 & 87.34 & 8.22 & 2.17 & 0.48 & 0.18 \\
    BST~\citep{smith2020bst} & 82.12 & 97.57 & 98.99 & 89.11 & 9.05 & 2.58 & 0.60 & 0.18 \\
    BS$\mathbb{B}$T (Ours) & 88.87 & 98.84 & 99.28 & 93.44 & 10.13 & 3.31 & 0.97 & 0.35 \\
    \midrule

    \textbf{Evaluation on ED benchmark} \\
    ConvAI2~\citep{zhang2018persona} & 47.90 & 76.14 & 85.87 & 60.60 & 8.66 & 1.79 & 0.60 & 0.32 \\
    WoW~\citep{dinan2019wizard} & 45.86 & 74.79 & 85.15 & 58.94 & 9.65 & 2.20 & 0.71 & 0.34 \\
    ED~\citep{rashkin2019ed} & 62.81 & 88.91 & 94.58 & 74.18 & 14.86 & 5.14 & 1.95 & 1.00 \\ 
    BST~\citep{smith2020bst} & 48.11 & 77.09 & 86.96 & 61.04 & 11.22 & 2.64 & 1.06 & 0.58 \\
    BS$\mathbb{B}$T (Ours) & 61.29 & 87.39 & 93.59 & 72.70 & 10.45 & 2.74 & 1.21 & 0.70 \\
    
    \bottomrule
\end{tabular}
}
\end{center}
\caption{Performance of retrieval and generative dialogue systems on all benchmarks used in this paper.}
\label{tab:evaluation_detail}
\end{table*}

\subsection{Moderator Agent}

In \textsc{BotsTalk} framework, the moderator agent leverages NLI classifier and skill classifier.

Given a response $res_{k,t}$ from a skill agent of skill $k$ and the set of skill contexts $\tilde{stx}$, the NLI classifier is designed to determine whether a response candidate contradicts any of the skill contexts. For NLI classifier, we employ the public HuggingFace\footnote{\url{https://github.com/huggingface}} implementation of a RoBERTa-Large model~\citep{liu2019roberta} fine-tuned on the Multi-Genre NLI dataset~\citep{adina2018mnli}. The RoBERTa model shows 90.59 accuracy on MNLI validation set. We regard each response candidate $res_k$ as a hypothesis sentence and each skill context $stx_k \in \tilde{stx}$ as a premise sentence, then conduct unidirectional NLI classification between $stx_k$ and $res_k$, determining whether a hypothesis sentence $res_k$ can be inferred from the given premise sentence $stx_k$ for all response candidates.

Given a response $res_t$, the skill classifier identifies the skill of the response among all skills represented in the skill context set $\tilde{stx}$. For skill classifier, we employ a BERT-base~\citep{devlin2018bert} model. We trained the model on utterances from ConvAI2, WoW, ED train sets and their corresponding skills $k$ as labels. The model was trained with a batch size of 16, a learning rate of 2e-5 and epoch 3. The BERT model shows 81.95 accuracy on utterances from ConvAI2, WoW, ED test sets.

\subsection{Skill Context Retrieval}\label{sec:skill}

In Appendix~\ref{sec:single}, we explore how we define the skill context for each skill considering the original settings from single-skill dialogue datasets. We now describe how we construct the seed information (\eg, skill contexts for each of skil types and dialogue context of 2 turns) for skill agents to start a conversation. We first collect consecutive utterance pairs from ConvAI2, WoW, and ED as seed utterance pairs and define it as $dtx$. We then follow the convention of past research, which inject a target communicative skill to dialogue systems by providing an extra description about the specific skill, \ie, skill context. As we aim to build a multi-skill chatbot, there is a need for integrating different dialogue setups from multiple single-skill datasets. For a generalizable dialogue setup, we retrieve relevant skill context for each skill by querying $dtx$ which is the seed utterance pair. Here, to gather seed information as much as possible, we match top-5 relevant skill contexts per utterance pair, which we give us five times the seed information. We use TF-IDF~\citep{chen2017reading} to find relevant skill contexts and a SQLite database for storing the sparse TF-IDF matrix. Note that we use a simple IR baseline as a lower bound since it is not our main focus. One can easily try other IR systems for more sophisticated setting.

\section{Implementation Details of Baselines}\label{sec:baselines}

We conduct a set of experiments with retrieval and generative tasks to cover diverse baselines. We provide training details of these baseline models.

\vspace{1mm}
\noindent\textbf{Retrieval Task} We adopt a 256-million parameter poly-encoder~\citep{humeau2019poly} pre-trained on pushshift.io Reddit dataset as a base architecture for the retrieval task. We fine-tune this base architecture on individual datasets, \ie, ConvAI2, WoW, ED, BST, and BS$\mathbb{B}$T for 8 epochs with batch size 128 and learning rate 5e-5. The learning rate decays by a factor of 0.4 upon plateau of the loss evaluated on the valid set every half epoch. The best parameters are chosen based on R@1 score.

\vspace{1mm}
\noindent\textbf{Generative Task} As a base architecture for the generative task, we adopt a pre-trained BART~\citep{lewis2020bart} with 12 layers in each of the encoder and decoder. We fine-tune this base architecture on individual datasets, \ie, ConvAI2, WoW, ED, BST, and BS$\mathbb{B}$T for 3 epochs with batch size 32 and learning rate 1.0. The learning rate decays by a factor of 0.3 upon plateau of the loss evaluated on the valid set every half epoch. The best parameters are chosen based on accuracy score. We use greedy sampling as a decoding strategy.

\begin{figure*}[t!]
\centering
    \includegraphics[width=130mm]{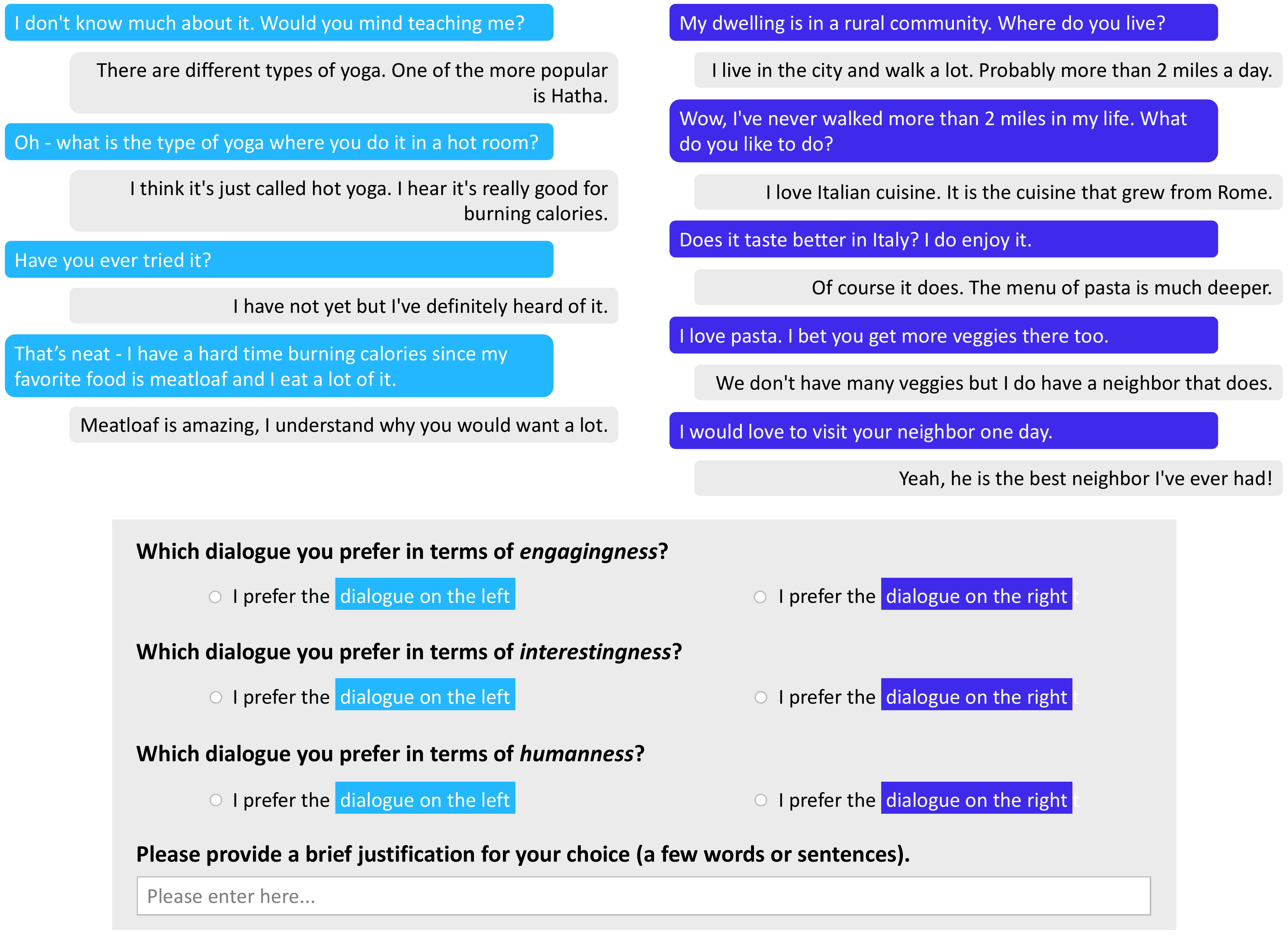}
\caption{Interface for human evaluation, which provides a pair of dialogues from BST (left) and BS$\mathbb{B}$T (right).}
\label{fig:bst_bsbt_pairwise}
\end{figure*}

\section{Additional Performance}\label{sec:additional}

We provide the evaluation results on all dialogue benchmarks used in this paper, \ie, BST benchmark for multi-skill benchmark, and ConvAI2, WoW, ED behcmarks for single-skill benchmarks. We report R@1, R@5, R@10, MRR for evaluating retrieval dialogue models, and BLEU-1, BLEU-2, BLEU-3, BLEU-4 for evaluating generative dialogue models. Table~\ref{tab:evaluation_detail} presents the performance of all of the baselines on all dialogue benchmarks.

\section{Settings of Human Evaluation}\label{sec:human}

We describe the specific settings for human evaluation that we perform to validate the quality of BS$\mathbb{B}$T dataset. Specifically, we employ ACUTE-Eval~\citep{li2019acute}, a widely used metric for multi-turn dialogue evaluation~\citep{dinan2020queens,li2020don}. We randomly sample 100 dialogues from BST and BS$\mathbb{B}$T datasets respectively. We only include dialogue contexts and exclude skill contexts for anonymity, since the skill contexts of BST and BS$\mathbb{B}$T are distinguishable. Figure~\ref{fig:bst_bsbt_pairwise} shows the interface used for human evaluation. We ask judges to compare each pair of conversations over three axes: engagingness, interestingness and humanness. 
The wording of questions is presented as follows:

\begin{itemize}
    \item \textbf{Engagingness:} Who would you prefer to talk to? Which version is more likely to hold your attention and make you want to hear more?
    \item \textbf{Interestingness:} Who would you say is more interesting? Which version arouses your curiosity or tells you something new or useful?
    \item \textbf{Humanness:} Who would you say sounds more human? Which version is more natural and personable?
\end{itemize}

\section{Qualitative Analysis}\label{sec:qualitative}

In what follows we conduct qualitative analysis on BS$\mathbb{B}$T dataset based on diverse samples. In each dialogue episode from BS$\mathbb{B}$T, one speaker is given five personas as $stx_\text{P}$, one topic as $stx_\text{K}$, and a situation description and emotion as $stx_\text{E}$, while another speaker is given five personas as $stx_\text{P}$, the topic and seven knowledge resources as $stx_\text{K}$, and nothing for $stx_\text{E}$. Each speaker is conditioned on their corresponding set of skill contexts, and annotates the response turn by turn. To give a deeper analysis, we select best-case (\ie, cherry-picked) and worst-case (\ie, lemon-picked) examples from BS$\mathbb{B}$T dataset. Table~\ref{tab:bsbt_cherry} and Table~\ref{tab:bsbt_lemon} shows cherry-picked examples and lemon-picked examples, respectively. We also present more dialogue examples randomly sampled from BS$\mathbb{B}$T in Table~\ref{tab:bsbt_random1} and Table~\ref{tab:bsbt_random2}.

\vspace{1mm}
\noindent\textbf{Cherry-picked Dialogues} In Table~\ref{tab:bsbt_cherry}, we provide cherry-picked dialogue samples from the BS$\mathbb{B}$T dataset. Example 1 and 2 both show that the conversation contains a sufficient amount of utterances to learn skill grounding for skill type P and E. In Example 3, each skill type is sustained for more than 2 turns, which allows dialogue systems to learn the ability of skill grounding. Example 3 also demonstrates all of the skill types are blended in a natural and reasonable way, which enables dialogue systems to weave multiple skills in a single conversation properly.

\vspace{1mm}
\noindent\textbf{Lemon-picked Dialogues} In Table~\ref{tab:bsbt_lemon}, we present lemon-picked conversations from BS$\mathbb{B}$T dataset, which fail to satisfy the key criteria of multi-skill dialogues. In Example 4, speaker A and B repeats greeting such as ``\emph{You as well. Have a good weekend as well}'' and ``\emph{You too. Have a good weekend as well}'' for about 3 turns. This is mainly because we set conversations to be of fixed length (\ie, 10 turns), while the conversation may end earlier than the given turns. To prevent these error cases, it will be helpful to have termination condition of conversations. We leave this for future work. Example 5 only features one skill type, which we already observe a few cases in Figure~\ref{fig:skill_count}. While the conversation may make sense, lack of skill blending in the conversation prevents a dialogue model from learning multiple skills. Meanwhile, it is widely known that dialogue agents suffer from generic responses~\citep{li2016persona}, as shown in Example 6. We observe that while shifts between skill are frequent, responses tend to show little relevance with the skill contexts of their corresponding skills. They often rephrase or repeat what has been discussed during conversation.

\begin{table*}[t!]
\begin{center}\small
{
\begin{tabular}{l | l}
    \toprule
    \multicolumn{2}{l}{\textbf{Example 1}} \\
    \midrule
    \textbf{Persona 1} & \textbf{Persona 2} \\
    I like to ski & I am an artist \\
    My wife does not like me anymore & I have four children \\
    I have went to Mexico 4 times this year & I recently got a cat \\
    I hate Mexican food & I enjoy walking for exercise \\
    I like to eat cheetos & I love watching Game of Thrones \\
    \midrule
    \multicolumn{2}{l}{PERSON1: Hi} \\
    \multicolumn{2}{l}{PERSON2: Hello ! How are you today?} \\
    \multicolumn{2}{l}{PERSON1: I am good thank you, how are you.} \\
    \multicolumn{2}{l}{PERSON2: Great, thanks ! My children and I were just about to watch Game of Thrones.} \\ 
    \multicolumn{2}{l}{PERSON1: Nice ! How old are your children?} \\
    \multicolumn{2}{l}{PERSON2: I have four that range in age from 10 to 21. You?} \\
    \multicolumn{2}{l}{PERSON1: I do not have children at the moment.} \\
    \multicolumn{2}{l}{PERSON2: That just means you get to keep all the popcorn for yourself.} \\
    \multicolumn{2}{l}{PERSON1: And Cheetos at the moment!} \\
    \multicolumn{2}{l}{PERSON2: Good choice. Do you watch Game of Thrones?} \\
    \multicolumn{2}{l}{PERSON1: No, I do not have much time for TV.} \\
    \multicolumn{2}{l}{PERSON2: I usually spend my time painting: but, I love the show.} \\
    
    \midrule
    \multicolumn{2}{l}{\textbf{Example 2}} \\
    \midrule
    \textbf{Persona 1} & \textbf{Persona 2} \\
    I enjoy going to museums & I like to party \\
    I am married & My major is business \\
    I live in New york city & I am in college \\
    I have ten nieces and nephews & I love the beach \\
    I am a writer & I work part time at a pizza restaurant \\
    \midrule
    \multicolumn{2}{l}{PERSON1: Hi. How are you doing today?} \\
    \multicolumn{2}{l}{PERSON2: Hi, I am great just finishing up some homework. How are you?} \\
    \multicolumn{2}{l}{PERSON1: I am alright. I just got done writing.} \\
    \multicolumn{2}{l}{PERSON2: Do you write for a living or as a hobby?} \\ 
    \multicolumn{2}{l}{PERSON1: It is my living. I like culture.} \\
    \multicolumn{2}{l}{PERSON2: That sounds like a fun job. I am a business major but have a part time job.} \\
    \multicolumn{2}{l}{PERSON1: What are you going to school for?} \\
    \multicolumn{2}{l}{PERSON2: I am trying to get my ba in finance.} \\
    \multicolumn{2}{l}{PERSON1: Do you own your own company?} \\
    \multicolumn{2}{l}{PERSON2: No still in school work at pizza hut part time.} \\
    \multicolumn{2}{l}{PERSON1: Do you get free pizza?} \\
    \multicolumn{2}{l}{PERSON2: No but I do get a good deal.} \\
    
    \midrule
    \multicolumn{2}{l}{\textbf{Example 3}} \\
    \midrule
    \textbf{Persona 1} & \textbf{Persona 2} \\
    I work at McDonald's & I want to be successful \\
    My parents did not want me & I have got two more years in college \\
    I was raised by my uncle and aunt & I have no siblings \\
    I am considering going to college & I am a student \\
    I am looking for something different this year & I study law \\
    \midrule
    \multicolumn{2}{l}{PERSON1: My mom had me in McDonald's bathroom when she was 12.} \\
    \multicolumn{2}{l}{PERSON2: That is really sad. Do you have any siblings?} \\
    \multicolumn{2}{l}{PERSON1: No. My uncle is the manager and took me home.} \\
    \multicolumn{2}{l}{PERSON2: That was nice of him. I am almost done with college.} \\
    \multicolumn{2}{l}{PERSON1: He gave me a job at McDonalds. I was considering college but I am good now.} \\
    \multicolumn{2}{l}{PERSON2: I am almost done. I only have two years left in law school.} \\
    \multicolumn{2}{l}{PERSON1: You be smart. I would like to do something different but I team leader now.} \\
    \multicolumn{2}{l}{PERSON2: That sounds important. Maybe one day you will go back to school.} \\
    \multicolumn{2}{l}{PERSON1: I dropped out to work. Maybe you can help me sure my mom for abandoning me?} \\
    \multicolumn{2}{l}{PERSON2: I do not know if we would be able to do anything because is has been so long.} \\
    \multicolumn{2}{l}{PERSON1: Oh I am only 15 right now. Is that too old?} \\
    \multicolumn{2}{l}{PERSON2: Do not you have to be older than 15 to work at McDonald's? That sounds illegal.} \\
    \bottomrule
\end{tabular}}
\end{center}
\caption{3 sample conversations from ConvAI2 dataset. In dialogue episodes, each speaker is given five persona sentences respectively.}
\label{tab:conv_dialogue_example}
\end{table*}

\begin{table*}[t!]
\begin{center}\small
{
\begin{tabular}{l l}
    \toprule
    \multicolumn{2}{l}{\textbf{Example 1}} \\
    \midrule
    \textbf{Topic:} & Armadillo \\
    \cmidrule{2-2}
    Wizard: & I love animals and think armadillos are awesome with their leathery shell. \\ 
    Apprentice: & I don’t think I’ve ever seen an armadillo in real life!\\
    Wizard: & I’ve seen them at the zoo. Armadillo means little armored one in Spanish. \\
    Apprentice: & Are they native to a Spanish-speaking part of the world? \\
    \cmidrule{2-2}
    \textbf{Knowledge:} & Armadillos are New World placental mammals in the order Cingulata ... \\
    & The word “armadillo” means “little armoured one” in Spanish. \\
    & It is usually made from dairy products, such as milk and cream, and ... \\
    & ... \\
    & The nine-banded armadillo (“Dasypus novemcinctus”), or the nine-banded, long-nosed \\
    & armadillo, is a medium-sized mammal found in North, Central, and South America. \\
    \cmidrule{2-2}
    Wizard: & Yes, they are most commonly found in North, Central, and South America \\ 
    \midrule
    
    \multicolumn{2}{l}{\textbf{Example 2}} \\
    \midrule
    \textbf{Topic:} & Ice cream \\
    \cmidrule{2-2}
    Wizard: & I just love ice cream. I love the types with fruits and flavours. Do you like ice cream? \\ 
    Apprentice: & I love Ice cream as much as any one. I especially like Gelato, foreign ice cream! \\
    \cmidrule{2-2}
    \textbf{Knowledge:} & Ice cream is a sweetened frozen food typically eaten as a snack or dessert ... \\
    & It is usually made from dairy products, such as milk and cream, and ... \\
    & ... \\
    & Bacon ice cream (or bacon-and-egg ice cream) is an ice cream generally created by \\
    & adding bacon to egg custard and freezing the mixture. \\
    \cmidrule{2-2}
    Wizard: & Me too. There are some strange combinations though, have you heard of bacon ice \\ 
    & cream? where they add bacon and even egg custard to the freezing mixture! \\ 
    Apprentice: & Surprisingly bacon ice cream doesn’t surprise me. That doesn’t sound appealing to me, \\
    & but perhaps it could be delicious ... \\
    
    \midrule
    \multicolumn{2}{l}{\textbf{Example 3}} \\
    \midrule
    \textbf{Topic:} & Lifeguard \\
    \cmidrule{2-2}
    Apprentice: & So I am a lifeguard. Know anything about saving lives in water? \\
    Wizard: & I’m impressed! It’s a big responsibility to supervise other people’s safety in the water! \\
    & Tell me more. \\
    Apprentice: & Well, I help make sure people do not drown or get injured while in or near the water! \\
    \cmidrule{2-2}
    \textbf{Knowledge:} & A lifeguard is a rescuer who supervises the safety and rescue of swimmers, surfers, ... \\
    & Lifeguards are strong swimmers and trained in CPR/AED first aid, certified in water ... \\
    & ... \\
    & In some areas, the lifeguard service also carries out mountain rescues, or may function \\
    & as the primary EMS provider. \\
    \cmidrule{2-2}
    Wizard: & I’ve heard that in some places, lifeguards also help with other sorts of emergencies, like \\
    & mountain rescues! Is that part of your job too? \\
    Apprentice: & I have! I feel like you know much about this! What brings you to know so much? \\
    Wizard: & Oh, that’s about the extent of my knowledge. I’ve just been around beaches and I’ve \\ 
    & always admired lifeguards. I’m not a super strong swimmer myself. \\
    \bottomrule
\end{tabular}
}
\end{center}
\caption{3 sample conversations from Wizard of Wikipedia dataset. In each dialogue episode, apprentice is given a topic, while wizard is given the same topic and access to an information retrieval system over Wikipedia. For each utterance, knowledge retrieval is performed based on dialogue history, giving about 61 knowledge candidates per turn.}
\label{tab:wow_dialogue_example}
\end{table*}

\begin{table*}[t!]
\begin{center}\small
{
\begin{tabular}{l l}
    \toprule
    \multicolumn{2}{l}{\textbf{Example 1}} \\
    \midrule
    \textbf{Emotion:} & Terrified (label) \\
    \textbf{Situation:} & My brother jump scared me while I was out playing. It was crazy bad. \\
    \midrule
    Speaker: & Just got scared to death. \\
    Listener: & Oh no. What happened? \\
    Speaker: & My brother jumped scared me. \\
    Listener: & lol is he younger or older? \\

    \toprule
    \multicolumn{2}{l}{\textbf{Example 2}} \\
    \midrule
    \textbf{Emotion:} & Proud (label) \\
    \textbf{Situation:} & My little dog learned to sit! \\
    \midrule
    Speaker: & I finally tough my new little puppy his first trick! \\
    Listener: & What trick did you teach him? \\
    Speaker: & I tought him to sit for a treat, it's so cute. \\
    Listener: & That is good, do you plan to teach him more tricks? \\
    
    \toprule
    \multicolumn{2}{l}{\textbf{Example 3}} \\
    \midrule
    \textbf{Emotion:} & Apprehensive (label) \\
    \textbf{Situation:} & I have to call my landlord about being late on the rent. I really don’t want to have this conversation. \\
    \midrule
    Speaker: & I have to make a dreadful phone call tomorrow. \\
    Listener: & Oh no, about what? \\
    Speaker: & I’m late on my rent and I need another week. I don’t want to because my landlord isnt very nice. \\
    Listener: & Oh no, I’ve been there done that too many times. \\
    Speaker: & I don’t want her to make a big deal. \\
    
    \toprule
    \multicolumn{2}{l}{\textbf{Example 4}} \\
    \midrule
    \textbf{Emotion:} & Content (label) \\
    \textbf{Situation:} & Eating my favorite meal makes me happy. \\
    \midrule
    Speaker: & I am at my best when I have my favorite meal. \\
    Listener: & Nice. \\
    Speaker: & I love enchiladas. \\
    Listener: & Really? \\

    \toprule
    \multicolumn{2}{l}{\textbf{Example 5}} \\
    \midrule
    \textbf{Emotion:} & Joyful (label) \\
    \textbf{Situation:} & I have had a great week! \\
    \midrule
    Speaker: & I have had a great start to my week! \\
    Listener: & That’s great. Do you think the rest of the week will be as great? \\
    Speaker: & I hope so! It looks promising!! \\
    Listener: & Lucky you. Are you always a positive person or it’s just been an amazing week really? \\
    
    \toprule
    \multicolumn{2}{l}{\textbf{Example 6}} \\
    \midrule
    \textbf{Emotion:} & Proud (label) \\
    \textbf{Situation:} & I was proud when my brother finished college. He worked so hard at it. \\
    \midrule
    Speaker: & I was proud of my brother when he finished school. He worked so hard at it. \\
    Listener: & Nice, tell him congrats. What did he major in? \\
    Speaker: & It was English. \\
    Listener: & He should become an English teacher. \\
    \bottomrule
\end{tabular}
}
\end{center}
\caption{6 sample conversations from Empathetic Dialogues dataset. In each dialogue episode, speaker is given a situation with an emotion, while listener is given nothing as input.}
\label{tab:ed_dialogue_example}
\end{table*}

\begin{table*}[t!]
\begin{center}\small
{\begin{tabular}{l | l}
    \toprule
    \multicolumn{2}{l}{\textbf{Example 1}} \\
    \midrule
    \textbf{Skill context from ConvAI2} & \textbf{Skill context from ConvAI2} \\
    I like to make cookies; I like bagels with everything on; ... & I love the smell of beans; ... \\
    \textbf{Skill context from WoW} & \textbf{Skill context from WoW} \\
    Obesity in the United States & Obesity in the United States; Obesity in the ...\\
    \textbf{Skill context from ED} & \textbf{Skill context from ED}  \\
    My girlfriend told me she's pregnant; Surprised & None \\
    \midrule
    \multicolumn{2}{l}{\textbf{Dialogue}} \\
    \multicolumn{2}{l}{A: Oh wow what kinds of videos do you have? (P)} \\ 
    \multicolumn{2}{l}{B: I do mainly make up tutorials. Do you have any side hobbies? (P)} \\
    \multicolumn{2}{l}{A: I like making cookies and eating cookie dough, I love bagels too. (P)} \\
    \multicolumn{2}{l}{B: I love bagels as well! What else do you like to do? (P)} \\
    \multicolumn{2}{l}{A: I also love playing sports, but I'm too busy at this moment. (P)} \\
    \multicolumn{2}{l}{B: You should try joining a fitness team! It will make you feel better about you being busy. (E)} \\   
    \multicolumn{2}{l}{A: LoL I'll, but I'd feel like I wouldn't be ready. (P)} \\
    \multicolumn{2}{l}{B: You could always try it out! It's always fun to try out new things! (E)} \\  
    \multicolumn{2}{l}{A: I would if I could I really appreciate new things and learn new things from people like you. (E)} \\
    \multicolumn{2}{l}{B: I wish you could someday. Thanks for chatting. (E)} \\ 
    \midrule
    \multicolumn{2}{l}{\textbf{Example 2}} \\
    
    \midrule
    \textbf{Skill context from ConvAI2} & \textbf{Skill context from ConvAI2} \\
    I like to read books; I can play the piano; ... & Halloween is my favorite holiday; ... \\
    \textbf{Skill context from WoW} & \textbf{Skill context from WoW} \\
    Vitamin C & Vitamin C; Vitamin C is a vitamin found in food ... \\
    \textbf{Skill context from ED} & \textbf{Skill context from ED}  \\
    I love walking my dogs on late summer mornings ...; Content & None \\
    \midrule
    \multicolumn{2}{l}{\textbf{Dialogue}} \\
    \multicolumn{2}{l}{A: I'm not nearly as energetic. My dogs take up most of my time unfortunately. (E)} \\ 
    \multicolumn{2}{l}{B: Well, at least you take care of them. (E) } \\
    \multicolumn{2}{l}{A: I don't mind them since they're my dogs. Just wish they were more active. (E)} \\
    \multicolumn{2}{l}{B: Well, what do you like to do to stay active? (P)} \\ 
    \multicolumn{2}{l}{A: I play the piano as my hobby! Are you interested in basketball? (P) } \\
    \multicolumn{2}{l}{B: Of course. I work a lot so basketball is important. (P) } \\   
    \multicolumn{2}{l}{A: Do you like to read? I love to read about chinese history. (P) } \\
    \multicolumn{2}{l}{B: I do but mostly I read about halloween on the weekends. (P)} \\  
    \multicolumn{2}{l}{A: That's fun! Halloween is such a fun holiday. (E) } \\
    \multicolumn{2}{l}{B: Yes I agree. (E) } \\
    \midrule
    
    \multicolumn{2}{l}{\textbf{Example 3}} \\
    \midrule
    \textbf{Skill context from ConvAI2} & \textbf{Skill context from ConvAI2} \\
    I enjoy skiing; My favorite music is rock and roll; ... & I've started working as a freelance designer;  ... \\
    \textbf{Skill context from WoW} & \textbf{Skill context from WoW} \\
    Pecan pie & Pecan pie; Pecan pie is a pie of pecan nuts mixed ... \\
    \textbf{Skill context from ED} & \textbf{Skill context from ED}  \\
     I wanted to do something to show my Mom some ... ; Caring & None \\
    \midrule

    \multicolumn{2}{l}{\textbf{Dialogue}} \\
    \multicolumn{2}{l}{A: Amazing to think some places divorce is illegal. I shudder to think what it must be like to be trapped like that. (K)} \\ 
    \multicolumn{2}{l}{B: At least in the US the courts can step in and help figure out all the child custody, support, and visitation issues. (K)} \\
    \multicolumn{2}{l}{A: Yeah, I agree. It's probably hard to make sure kids are all kept by both parents. (E)} \\
    \multicolumn{2}{l}{B: I'm sure that's certainly a big part of it, too. (E)} \\ 
    \multicolumn{2}{l}{A: Do you travel often? I really love skiing, but am considering going to the us for business. (P)} \\
    \multicolumn{2}{l}{B: I really like Canada. I'm thinking about going there for a business trip. (P)} \\
    \multicolumn{2}{l}{A: Canada is very beautiful ! Do you like music, too? I really enjoy rock and roll. (P) } \\
    \multicolumn{2}{l}{B: Yeah, I really love music. It's one of my hobbies. (P)} \\
    \multicolumn{2}{l}{A: That is good. What do you like to do for fun? (P)} \\
    \multicolumn{2}{l}{B: I like to draw, and be a good animator. You? (P)} \\
    \bottomrule
\end{tabular}}
\end{center}
\caption{Cherry-picked dialogue examples from BS$\mathbb{B}$T dataset.}
\label{tab:bsbt_cherry}
\end{table*}

\begin{table*}[t!]
\begin{center}\small
{\begin{tabular}{l | l}
    \toprule
    \multicolumn{2}{l}{\textbf{Example 4}} \\
    \midrule
    \textbf{Skill context from ConvAI2} & \textbf{Skill context from ConvAI2} \\
    I love spicy food; I like swimming; I am an only child; ... & My favorite food is jello; I cannot whistle; ... \\
    \textbf{Skill context from WoW} & \textbf{Skill context from WoW} \\
    Marathon & Marathon; The marathon is a long-distance running ...\\
    \textbf{Skill context from ED} & \textbf{Skill context from ED}  \\
    At work sometimes I put a lot of faith ... ; Trusting & None \\
    \midrule
    \multicolumn{2}{l}{\textbf{Dialogue}} \\
    \multicolumn{2}{l}{A: Lucky. I've to get things ready for tonights group event. (P) } \\ 
    \multicolumn{2}{l}{B: What's your event that you're doing? (P)} \\
    \multicolumn{2}{l}{A: Weekly conference where I can talk to people from my field. (E)} \\
    \multicolumn{2}{l}{B: That sounds like a lot of fun. Is this your first one? (E)} \\ 
    \multicolumn{2}{l}{A: This is the first one of my own. I hope to make it to the next conference. (E)} \\
    \multicolumn{2}{l}{B: Good luck with that. I hope you have fun. (E)} \\   
    \multicolumn{2}{l}{A: Thank you. I hope you have fun as well. (E)} \\
    \multicolumn{2}{l}{B: I hope you have a great weeked and good luck. (E)} \\  
    \multicolumn{2}{l}{A: You as well. Have a good weekend as well. (K)} \\
    \multicolumn{2}{l}{B: You too. Have a good weekend as well. (K)} \\
    \midrule
    
    \multicolumn{2}{l}{\textbf{Example 5}} \\
    \midrule
    \textbf{Skill context from ConvAI2} & \textbf{Skill context from ConvAI2} \\
    My friends and I watch the games together; We drink beer; ... & I like to swim; My favorite drink is chocolate milk; ... \\
    \textbf{Skill context from WoW} & \textbf{Skill context from WoW} \\
    Veganism & Veganism; Veganism is both the practice of ... \\
    \textbf{Skill context from ED} & \textbf{Skill context from ED}  \\
    My neighbor's above ground swimming pool ...; Confident & None \\
    \midrule
    \multicolumn{2}{l}{\textbf{Dialogue}} \\
    \multicolumn{2}{l}{A: I meant actually swimming in the pool. (P)} \\ 
    \multicolumn{2}{l}{B: I really only swim in natural ponds (NSP) because I have an autoimmune disease. (P)} \\
    \multicolumn{2}{l}{A: Oh wow. Did you eat some bad foods? (P)} \\
    \multicolumn{2}{l}{B: Well I have a nasty bald beard which is a major problem. What do you do? (P)} \\ 
    \multicolumn{2}{l}{A: I'm an architect. I love the money I get from it. (P)} \\
    \multicolumn{2}{l}{B: What is your favorite dessert? I like tacos. (P)} \\   
    \multicolumn{2}{l}{A: I like chocolate so much. What is your favorite food? (P)} \\
    \multicolumn{2}{l}{B: I love all chocolate milk drinks. It is my favorite. (P)} \\  
    \multicolumn{2}{l}{A: I love chocolate too. Especially vanilla. You? (P)} \\
    \multicolumn{2}{l}{B: I love all the vanilla flavors. (P)} \\
    \midrule
    
    \multicolumn{2}{l}{\textbf{Example 6}} \\
    \midrule
    \textbf{Skill context from ConvAI2} & \textbf{Skill context from ConvAI2} \\
    I listen to podcasts; I like to read ... & I like makeup; many people want to be my friend; ... \\
    \textbf{Skill context from WoW} & \textbf{Skill context from WoW} \\
    Cake Decorating & Cake Decorating; Cake decorating is one of the ... \\
    \textbf{Skill context from ED} & \textbf{Skill context from ED}  \\
    Been making a lot of money off of my hobby lately ...; Surprised & None \\
    \midrule
    \multicolumn{2}{l}{\textbf{Dialogue}} \\
    \multicolumn{2}{l}{A: That sounds like an exciting hobby. (E)} \\ 
    \multicolumn{2}{l}{B: It is. Super fun and good exercise. What are your hobbies? (P)} \\
    \multicolumn{2}{l}{A: I like to listen to podcasts. My favorite is free piano. (P)} \\
    \multicolumn{2}{l}{B: I play the piano too. And makeup and exercise. I really enjoy those things. (P)} \\ 
    \multicolumn{2}{l}{A: It's hard to do fun things when you're not having fun. (E)} \\
    \multicolumn{2}{l}{B: Very true. I can be kind to my friends when I am having fun. (P)} \\   
    \multicolumn{2}{l}{A: I am sure your friends will be happy to have you as a friend. (K)} \\
    \multicolumn{2}{l}{B: They sure will. They want to be my friend too! (P)} \\  
    \multicolumn{2}{l}{A: That is so sweet of you! You have great friends! (K)} \\
    \multicolumn{2}{l}{B: They have a lot of a nice smile so I try to be nice. (P)} \\
    \bottomrule
    
\end{tabular}}
\end{center}
\caption{Lemon-picked dialogue examples from BS$\mathbb{B}$T dataset.}
\label{tab:bsbt_lemon}
\end{table*}

\begin{table*}[t!]
\begin{center}\small
{\begin{tabular}{l | l}
    \toprule
    \multicolumn{2}{l}{\textbf{Example 1}} \\
    \midrule
    \textbf{Skill context from ConvAI2} & \textbf{Skill context from ConvAI2} \\
    I like to go for runs; I live by myself; ... & I like to listen to music; I have a boxer dog; ... \\
    \textbf{Skill context from WoW} & \textbf{Skill context from WoW} \\
    Yoga & Yoga; Yoga is a group of practices or disciplines ...\\
    \textbf{Skill context from ED} & \textbf{Skill context from ED}  \\
    I felt really guilty when I had to leave my dog ...; Guilty & None \\
    \midrule
    \multicolumn{2}{l}{\textbf{Dialogue}} \\
    \multicolumn{2}{l}{A: Hey, do you have any kids? (P)} \\ 
    \multicolumn{2}{l}{B: No. I shop in my spare time. Do you? (P)} \\
    \multicolumn{2}{l}{A: Yes, I do. Not the only shopper here. (P)} \\
    \multicolumn{2}{l}{B: Nice! Do you like to swim? I am all about that exercise. (P)} \\
    \multicolumn{2}{l}{A: I love swimming. I wish I could swim more though. (P)} \\
    \multicolumn{2}{l}{B: I wish I could swim too. I am also much older. (E)} \\   
    \multicolumn{2}{l}{A: I'm not as young as I used to be. I lost weight then. I am wondering how old you are. (E)} \\
    \multicolumn{2}{l}{B: I am 50 years old. Are you in school? (P)} \\  
    \multicolumn{2}{l}{A: Yep. I think its time to start. You don't like to wait long. (P)} \\
    \multicolumn{2}{l}{B: Definitely. That is the life stage for me. I hope I can get back to age. (E)} \\ 
    \midrule

    \multicolumn{2}{l}{\textbf{Example 2}} \\
    \midrule
    \textbf{Skill context from ConvAI2} & \textbf{Skill context from ConvAI2} \\
    I'm going on a cruise next month.; I love to cook.; ... & I like going barefoot.; I can knit.; ... \\
    \textbf{Skill context from WoW} & \textbf{Skill context from WoW} \\
    Cooking & Cooking; Cooking is the art of preparing food for ... \\
    \textbf{Skill context from ED} & \textbf{Skill context from ED}  \\
    I tried a new recipe the other day. It did not ...; Apprehensive & None \\
    \midrule
    \multicolumn{2}{l}{\textbf{Dialogue}} \\
    \multicolumn{2}{l}{A: Did you do any sight seeing when you last went to Dallas? (K)} \\ 
    \multicolumn{2}{l}{B: Not really, we were there for a friends wedding. Mind you, we did go to some great restaurants. With a population of} \\
    \multicolumn{2}{l}{$\quad$ over a million. People the city has a huge number of quality places to go! (K)} \\
    \multicolumn{2}{l}{A: I would have thought the same, but then again we have so many great dishes to eat. Did you enjoy any of them? (P)} \\
    \multicolumn{2}{l}{B: I enjoyed them all! We stayed for more than one night, so I know how to do some great things! (P)} \\ 
    \multicolumn{2}{l}{A: I am excited because I'm going on a cruise. (E)} \\
    \multicolumn{2}{l}{B: I hope you have a great time! Cruise is such a great destination! (E)} \\   
    \multicolumn{2}{l}{A: There are so many different places to go on a cruise. Did you visit any other restaurants near where you were at? (P)} \\
    \multicolumn{2}{l}{B: There were a lot! I saw a restaurant on my cruise tour that I had been to before. (P)} \\  
    \multicolumn{2}{l}{A: Was it expensive? Did you have a special restaurant? (P)} \\
    \multicolumn{2}{l}{B: No, it wasn't expensive at all. We had hot dogs and rice, which I enjoyed most of the time. (P)} \\ 
    \midrule

    \multicolumn{2}{l}{\textbf{Example 3}} \\
    \midrule
    \textbf{Skill context from ConvAI2} & \textbf{Skill context from ConvAI2} \\
    I go to high school.; I play ice hockey; ... &  I have a wife and two kids; I am a vegetarian; ... \\
    \textbf{Skill context from WoW} & \textbf{Skill context from WoW} \\
     Laziness & Laziness; Laziness is disinclination to activity or ... \\
    \textbf{Skill context from ED} & \textbf{Skill context from ED}  \\
    I participated in my high school runners' marathon ...; Proud & None \\
    \midrule
    \multicolumn{2}{l}{\textbf{Dialogue}} \\
    \multicolumn{2}{l}{A: One job would be better then my three. (P)} \\ 
    \multicolumn{2}{l}{B: What do you like to eat? I'm vegetarian. (P)} \\
    \multicolumn{2}{l}{A: I love beef broth, grilled cheese and eggs. (P)} \\
    \multicolumn{2}{l}{B: Yum! My wife and 2 kids do the same. (P)} \\ 
    \multicolumn{2}{l}{A: Family is also my life style. Maybe we should have a child! (E)} \\
    \multicolumn{2}{l}{B: But what if you didn't want a child? (E)} \\   
    \multicolumn{2}{l}{A: If I didn't want a child I probably wouldn't have them. (E)} \\
    \multicolumn{2}{l}{B: That's true! Our kids would just be us. (E)} \\ 
    \multicolumn{2}{l}{A: That's what I thought. (E)} \\
    \multicolumn{2}{l}{B: I understand that. (E)} \\
    \bottomrule
\end{tabular}}
\end{center}
\caption{Randomly-sampled dialogue examples from BS$\mathbb{B}$T dataset.}
\label{tab:bsbt_random1}
\end{table*}

\begin{table*}[t!]
\begin{center}\small
{\begin{tabular}{l | l}
    \toprule
    \multicolumn{2}{l}{\textbf{Example 4}} \\
    \midrule
    \textbf{Skill context from ConvAI2} & \textbf{Skill context from ConvAI2} \\
    I drive an SUV; I've 3 dogs; I have a law degree; ... & I've 3 small dogs; I tutor kids in my spare time; ... \\
    \textbf{Skill context from WoW} & \textbf{Skill context from WoW} \\
    Mexico City & Mexico City; Mexico City is the capital and most ... \\
    \textbf{Skill context from ED} & \textbf{Skill context from ED}  \\
    New apartment, family and friends I love, this is ...; Content & None \\
    \midrule
    \multicolumn{2}{l}{\textbf{Dialogue}} \\
    \multicolumn{2}{l}{A: How modern is your capital city? (K)} \\ 
    \multicolumn{2}{l}{B: It's been around a minute, capital and most populous city is Algiers, it's located in the far north of the country. (K)} \\
    \multicolumn{2}{l}{A: Very cool. How do you feel about it? Do you like your city or your country? (P)} \\
    \multicolumn{2}{l}{B: I love it and love the people. What do you do for a living? (P)} \\
    \multicolumn{2}{l}{A: I currently have a B.A. in law. Do you have any pets? (P)} \\
    \multicolumn{2}{l}{B: I have 3 little ones, and 2 girls. (P)} \\   
    \multicolumn{2}{l}{A: Nice! Will you move? (P)} \\
    \multicolumn{2}{l}{B: Yes, I would. The cost of living in Algiers is so high that I do not know if I would be able to afford it. (P)} \\  
    \multicolumn{2}{l}{A: It is expensive. Do you ever vacation? I like to visit the mountains. (P)} \\
    \multicolumn{2}{l}{B: Yes. I visit a lot of places there. (P)} \\
    \midrule
    
    \multicolumn{2}{l}{\textbf{Example 5}} \\
    \midrule
    \textbf{Skill context from ConvAI2} & \textbf{Skill context from ConvAI2} \\
    My favorite band is Bon Iver; I am from Seattle; ... & I like to attend wine tours; I enjoy visiting museums; ... \\
    \textbf{Skill context from WoW} & \textbf{Skill context from WoW} \\
    Marduk (band) & Marduk (band); Marduk is a black metal band from ... \\
    \textbf{Skill context from ED} & \textbf{Skill context from ED}  \\
    My best friend's husband cheated on her with ...; Disgusted & None \\
    \midrule
    \multicolumn{2}{l}{\textbf{Dialogue}} \\
    \multicolumn{2}{l}{A: I have never heard of a chupacabra. Would you like to live forever? (P)} \\ 
    \multicolumn{2}{l}{B: No. However I would like to listen to Oasis' hit "Live Forever" since you mention the idea. (K)} \\
    \multicolumn{2}{l}{A: That is true, I was hoping I could be of some help. (E)} \\
    \multicolumn{2}{l}{B: Anything that helps would be great. Thank you very much. (E)} \\
    \multicolumn{2}{l}{A: You're welcome. There is also a band called CTCR that plays a bit of rap as well. (K)} \\
    \multicolumn{2}{l}{B: I'll have to give them a listen. (E)} \\   
    \multicolumn{2}{l}{A: There is an album you should check out. It's called the on the court, you should really listen to it. (K)} \\
    \multicolumn{2}{l}{B: I'll have to take a listen to it. (E)} \\  
    \multicolumn{2}{l}{A: I have to go now, but I hope you can tell them how much you love it and how you felt in this situation. (E)} \\
    \multicolumn{2}{l}{B: Yes, I will do that as soon as possible. (E)} \\
    \midrule
    
    \multicolumn{2}{l}{\textbf{Example 6}} \\
    \midrule
    \textbf{Skill context from ConvAI2} & \textbf{Skill context from ConvAI2} \\
    I love coffee with cream; I went to New York city once; ... & I tutor kids in my spare time; I am very religious; ... \\
    \textbf{Skill context from WoW} & \textbf{Skill context from WoW} \\
    Dog & Dog; The domestic dog is a member of "Canis", ... \\
    \textbf{Skill context from ED} & \textbf{Skill context from ED} \\
    We were living in Hawaii ...; Content  & None \\
    \midrule
    \multicolumn{2}{l}{\textbf{Dialogue}} \\
    \multicolumn{2}{l}{A: That is funny. What does jaja mean? (K)} \\ 
    \multicolumn{2}{l}{B: Same as ll but in Spanish! (K)} \\
    \multicolumn{2}{l}{A: Well, that is interesting, I guess. (E)} \\
    \multicolumn{2}{l}{B: Well it's used for many cultures, Spanish comes after. What we get from it is also used in most western countries. (K)} \\ 
    \multicolumn{2}{l}{A: Okay, that is cool. Thank you again, I am learning Spanish also! (P)} \\
    \multicolumn{2}{l}{B: I'm learning English as well! I'm so happy for you! (P)} \\   
    \multicolumn{2}{l}{A: Thanks, that is interesting. I like learning a foreign language. (P)} \\
    \multicolumn{2}{l}{B: That's a really cool skill to have. I bet you get to be learning all sorts of different languages. (E)} \\ 
    \multicolumn{2}{l}{A: I sure do, because it's an important skill. (E)} \\
    \multicolumn{2}{l}{B: Did you take it in high school? I know a lot of native Spanish speakers. (P)} \\
    \bottomrule
\end{tabular}}
\end{center}
\caption{Randomly-sampled dialogue examples from BS$\mathbb{B}$T dataset.}
\label{tab:bsbt_random2}
\end{table*}

\end{document}